\documentclass[10pt,twocolumn,letterpaper]{article}

\usepackage{cvpr}              

\usepackage{graphicx}
\usepackage{amsmath}
\usepackage{amssymb}
\usepackage{booktabs}
\usepackage{blindtext}
\usepackage{enumitem}
\usepackage{multirow}
\usepackage{bm}
\usepackage[dvipsnames]{xcolor}
\usepackage{nicefrac}
\usepackage{soul}

\usepackage[pagebackref,breaklinks,colorlinks]{hyperref}

\usepackage[capitalize]{cleveref}
\crefname{section}{Sec.}{Secs.}
\Crefname{section}{Section}{Sections}
\Crefname{table}{Table}{Tables}
\crefname{table}{Tab.}{Tabs.}

\newcommand{\partitle}[1]{{\noindent}{\textbf{#1}}.}
\newcommand*\diff{\mathop{}\!\mathrm{d}}

\newcommand\blfootnote[1]{%
	\begingroup
	\renewcommand\thefootnote{}\footnote{#1}%
	\addtocounter{footnote}{-1}%
	\endgroup
}

\newcommand{\zb}[1]{\textcolor[rgb]{0.4, .0, .8}{{[Ziqian: #1]}}}
\newcommand{\ks}[1]{\textcolor[rgb]{0.8, .2, .4}{{[Kripa: #1]}}}
\newcommand{\ft}[1]{\textcolor[rgb]{0.4, .8, .4}{{[Feitong: #1]}}}

\def\cvprPaperID{}
\def\confName{CVPR}
\def\confYear{2023}

\begin{document}

\title{Learning Personalized High Quality Volumetric Head Avatars\\ from Monocular RGB Videos}




\author{Ziqian Bai$^{1,2*}$ \quad
Feitong Tan$^{1}$ \quad
Zeng Huang$^{1}$ \quad
Kripasindhu Sarkar$^{1}$ \quad
Danhang Tang$^{1}$ \\
Di Qiu$^{1}$ \quad
Abhimitra Meka$^{1}$ \quad
Ruofei Du$^{1}$ \quad
Mingsong Dou$^{1}$ \quad
Sergio Orts-Escolano$^{1}$ \\
Rohit Pandey$^{1}$ \quad
Ping Tan$^{2}$ \quad
Thabo Beeler$^{1}$ \quad
Sean Fanello$^{1}$ \quad
Yinda Zhang$^{1}$ \\

$^{1}$ Google \qquad $^{2}$ Simon Fraser University
}

\twocolumn[{%
\renewcommand\twocolumn[1][]{#1}%
\maketitle
\vspace{-11.98mm}
\begin{center}
    \centering
    \includegraphics[width=1.0\linewidth, trim={0 0 0 0}, clip]{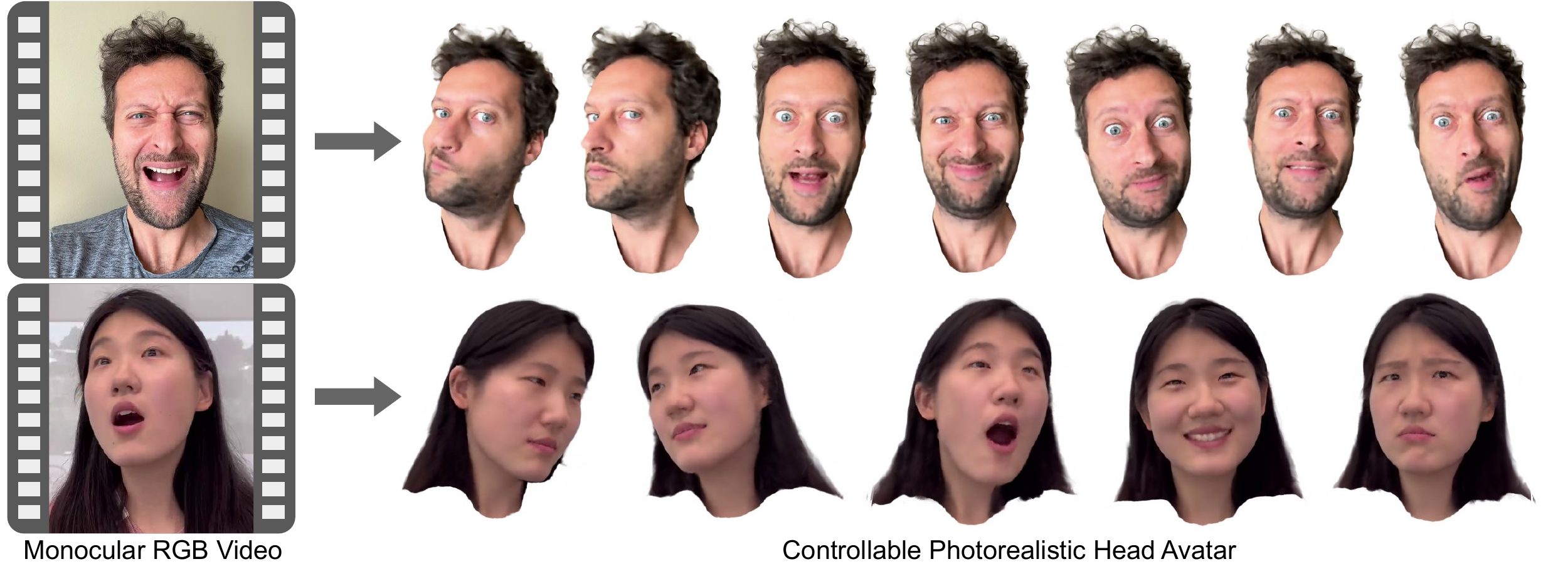}
    \vspace{-6mm}
    \captionof{figure}{
    Our technique builds a 3D avatar representation of a person using just a single short monocular RGB video (\eg, 1-2 minutes). We leverage a 3DMM to track the user's expressions. By anchoring a neural radiance field to the 3DMM geometry, we generate a volumetric photorealistic 3D avatar that can be rendered with user-defined expression and viewpoint. 
    Note that our method works well on challenging materials, \eg, hair and dramatic expressions. Please see our webpage \href{https://augmentedperception.github.io/monoavatar/}{augmentedperception.github.io/monoavatar} for more results.
    }
    \label{fig:teaser}
\end{center}%
}]

\begin{abstract}
    We propose a method to learn a high-quality implicit 3D head avatar from a monocular RGB video captured in the wild. The learnt avatar is driven by a parametric face model to achieve user-controlled facial expressions and head poses. Our hybrid pipeline combines the geometry prior and dynamic tracking of a 3DMM with a neural radiance field to achieve fine-grained control and photorealism. 
    To reduce over-smoothing and improve out-of-model expressions synthesis,
    we propose to predict local features anchored on the 3DMM geometry. These learnt features are driven by 3DMM deformation and interpolated in 3D space to yield the volumetric radiance at a designated query point. We further show that using a Convolutional Neural Network in the UV space is critical in incorporating spatial context and producing representative local features. Extensive experiments show that we are able to reconstruct high-quality avatars, with more accurate expression-dependent details, good generalization to out-of-training expressions, and quantitatively superior renderings compared to other state-of-the-art approaches.

\end{abstract}

\blfootnote{$^{*}$Work was conducted while Ziqian Bai was an intern at Google.}

\section{Introduction}

Creating a controllable human avatar is a fundamental piece of technology for many downstream applications, such as AR/VR communication~\cite{Orts-Escolano2016Holoportation,He2020CollaboVR}, virtual try-on~\cite{Saito2019PIFu}, virtual tourism~\cite{Du2019Geollery}, games~\cite{Waggoner2009My}, and visual effects for movies~\cite{Du2019Montage4D,Guo2019The}.
Prior art in high-quality avatar generation typically requires extensive hardware configurations (\ie, camera arrays~\cite{beeler2011high, Orts-Escolano2016Holoportation,Du2019Montage4D}, light stages~\cite{Guo2019The,Meka2020Deep}, dedicated depth sensors~\cite{Chen22}), or laborious manual intervention~\cite{MetaHuman2022Unreal}.
Alternatively, reconstructing avatars \textbf{from monocular RGB videos} significantly relaxes the dependency on equipment setup and broadens the application scenarios. However, monocular head avatar creation is highly ill-posed due to the dual problems of reconstructing and tracking highly articulated and deformable facial geometry, while modeling sophisticated facial appearance. 
Traditionally, 3D Morphable Models (3DMM)~\cite{blanz1999morphable,FLAME:SiggraphAsia2017} have been used to model facial geometry and appearance for various applications including avatar generation~\cite{ichim2015dynamic,garrido2016reconstruction,cao2016real}. However, 3DMMs do not fully capture subject-specific static details and dynamic variations, such as hair, glasses, and expression-dependent high frequency details such as wrinkles, due to the limited capacity of the underlying linear model.

Recent works \cite{athar2022rignerf,gafni2021dynamic} have incorporated neural radiance fields in combination with 3DMMs for head avatar generation to achieve photorealistic renderings, especially improving challenging areas, such as hair, and adding view-dependent effects, such as reflections on glasses. The pioneering work of NerFACE \cite{gafni2021dynamic} uses a neural radiance field parameterized by an MLP that is conditioned on 3DMM expression parameters and learnt per-frame latent codes.  While they achieve photorealistic renderings, the reliance on an MLP to directly decode from 3DMM parameter space leads to the loss of fine-grain control over geometry and articulation. 
Alternatively, RigNeRF \cite{athar2022rignerf} learns the radiance field in a canonical space by warping the target head geometry using a 3DMM fit, which is further corrected by a learnt dense deformation field parameterized by another MLP. While they demonstrate in-the-wild head pose and expression control, the use of two global MLPs to model canonical appearance and deformations for the full spatial-temporal space leads to a loss of high frequency details, and an overall uncanny appearance of the avatar. Both of these works introduce new capabilities but suffer from lack of detail in both appearance and motion because they attempt to model the  avatar's global appearance and deformation with an MLP network.

In this paper, we propose a method to learn a neural head avatar from a monocular RGB video. The avatar can be controlled by an underlying 3DMM model and deliver high-quality rendering of arbitrary facial expressions, head poses, and viewpoints, which retain fine-grained details and accurate articulations. We achieve this by learning to predict expression-dependent spatially local features on the surface of the 3DMM mesh.
A radiance field for any given 3D point in the volume is then obtained by interpolating the features from K-nearest neighbor vertices on the deformed 3DMM mesh in target expression, and passing them through a local MLP to infer density and color. The local features and local MLP are trained jointly by supervising the radiance field through standard volumetric rendering on the training sequence \cite{mildenhall2021nerf}. 
Note that our networks rely on the local features to model appearance and deformation details, and leverages the 3DMM to model only the global geometry. 

Learning local features is critical in achieving a high-quality head avatar. To this end, we train an image-to-image translation U-Net that transforms the 3DMM deformations in the UV space to such local features. These UV-space features are then attached to the corresponding vertices of the 3DMM mesh geometry. 
We show that learning features from such explicit per-vertex local displacement of the 3DMM geometry makes the model retain high-frequency expression-dependent details and also generalizes better to out-of-training expressions, presumably because of the spatial context between nearby vertices incorporated by the convolutional neural network (CNN).
An alternative approach is to feed the 3DMM parameters directly into a CNN decoder running on the UV space. However, we found this produces severe artifacts on out-of-training expressions, particularly given a limited amount of training data, \eg for a lightweight, 1-minute data collection procedure during the avatar generation process.

\if 
Another important, but often overlooked, issue in head avatar generation is the rendering of the mouth region. While theoretically 3DMMs are capable of modeling the mouth interior, practically, existing parametric models do not achieve this in a detailed manner, and would require significant new large scale data collection efforts to do so. We propose a simple approach to complement the mouth interior with derivable geometry on the existing 3DMM model, effectively improving the rendering quality and generalization to unseen expressions.
\fi

In summary, our contributions are as follows:
we propose a neural head avatar representation based on a 3DMM-anchored neural radiance field, which can model complex expression-dependent variations, but requires only monocular RGB videos for training.
We show that a convolutional neural network running on per-vertex displacement in UV space is effective in learning local expression-dependent features, and delivers favorable training stability and generalization to out-of-training expressions.
\if 
We further propose a simple yet effective approach to complement the mouth interior for 3DMMs, and hence improving the rendering quality.
\fi
Experiments on real-world datasets show that our model provides competitive controllability and generates sharper and detail enriched rendering compared to state-of-the-art approaches.


\if
Comparing to previous implicit representation methods, we use a different way to model out-of-model expression-dependent variations without deviating from the 3DMM constraints.
Prior works typically use 3DMM expression and pose parameters + MLPs (NerFACE)\zb{ for this purpose, which can be too abstract thus prone to overfit and poor in generalization. Following works are proposed to use 3DMM deformation + MLPs or learn corrected blendshapes in the form of implicit neural fields} (RigNeRF; IMAvatar). \ks{Can you check this claim? RigNeRF uses the 3DMM deformation to guide the NeRF -- in that case, it is more than expression and pose parameters + MLP.} \zb{However, they either do not consider 3DMM deformations of spatial neighbors during query, or have limited capacity to capture expression-dependent variations due to the linear formulation of blendshapes.}
We show that our design, \textit{i.e.}, vertex 3D deformations of expressions and poses + U-Net in UV space, can produce better rendering quality while still properly constraining the ill-posed monocular reconstruction with 3DMM priors.

While our model can capture out-of-model details on 3DMM, \eg hair and wrinkles on the face, rendering the mouth region correctly is still challenging presumably because of the poor capability of 3DMM in fitting the mouth region.
We xxxxx


In summary, our contributions are as follows.


\fi

\section{Related Works}
Building photorealistic representations of humans has been widely researched in the past few decades. Here, we mainly discuss prior art in head avatar and refer readers to the state-of-the-art surveys~\cite{zollhofer2018state,egger20203d} for a comprehensive literature review.

\begin{figure*}
    \centering
    \includegraphics[width=\linewidth]{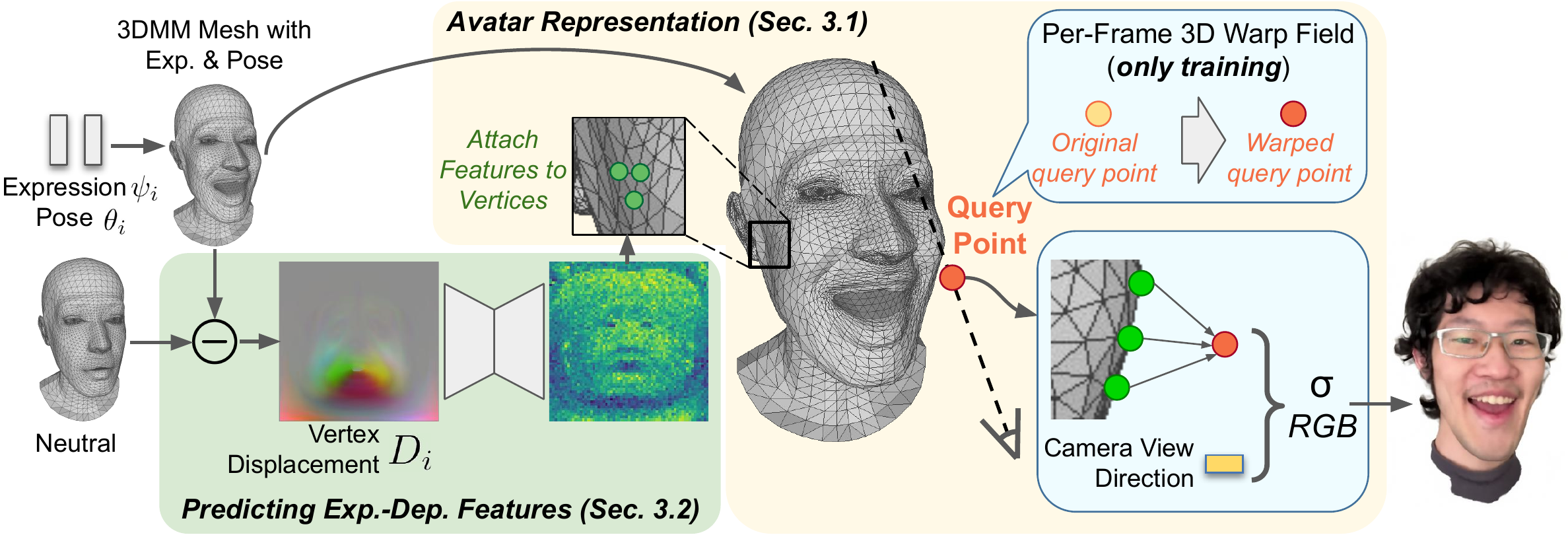}
    \vspace{-2em}
    \caption{Overview of our pipeline. The core of our method is the Avatar Representation (\cref{sec:avatar_representation}. Shown as the yellow area) based on a 3DMM-anchored neural radiance field (NeRF), which are decoded from local features attached on the 3DMM vertices. Then, we use volumetric rendering to compute the output image. To predict the vertex-attached features (\cref{sec:exp_dep_feats}. Shown as the green area), we first compute the vertex displacements from the 3DMM expression and pose, then process the displacements in UV space with Convolutional Neural Networks (CNNs), and sample the obtained features back to mesh vertices. 
    }
    
    \label{fig:overview}
\end{figure*}

\partitle{Monocular Explicit Surface Head (Face) Avatars}
Traditionally, a typical approach to create head (or face) avatars from monocular RGB videos is using a 3D Morphable Models (3DMM) as the foundation and adding personalized representations, such as corrected blendshapes~\cite{ichim2015dynamic,garrido2016reconstruction}, detail texture maps~\cite{ichim2015dynamic,garrido2016reconstruction}, image-based representations~\cite{cao2016real}, and secondary components~\cite{hu2017avatar}.
Early works use various optimizations to obtain the personalized representations from monocular data, including analysis-by-synthesis~\cite{zollhofer2018state,egger20203d,hu2017avatar,garrido2016reconstruction} as well as shape-from-shading~\cite{ichim2015dynamic,garrido2016reconstruction}.
Recent approaches replace optimizations by regressions with Deep Neural Networks (DNNs)~\cite{chaudhuri2020personalized,yang2020facescape,tewari2019fml}, or integrate optimizations with deep learning components~\cite{bai2020deep,Bai_2021_CVPR}. More recent methods leverage neural textures~\cite{grassal2022neural} to generate photorealistic appearances.

The main drawback of these methods is that they rely on explicit meshes with a fixed topology, making it hard to handle out-of-model details such as hair and accessories like glasses and apparels.
In contrast, our hybrid method combines geometric priors with implicit representations, leading to a significantly larger representation capacity.







\partitle{Monocular Implicit Head Avatars}
Recent work proposes to extend 3DMM with implicit 3D representations. NerFACE~\cite{gafni2021dynamic} introduces a dynamic neural radiance field (NeRF) conditioned on 3DMM expression codes which can render a view-consistent avatar with volumetric rendering. 
Since NerFACE directly inputs the 3DMM expression codes into MLPs without using any shape or spatial information from 3DMM, their model is quite under-constrained for monocular reconstruction, and suffers from severe artifacts for data with challenging expressions. 
RigNeRF~\cite{athar2022rignerf} uses 3DMM derived warping field to deform the camera space into a canonical space, and defines a canonical NeRF conditioned on 3DMM codes. However, their model uses a dense MLP-based architecture to memorize the appearance and deformation for the full head, leading to oversmooth results due to limited network capacity.  IMAvatar~\cite{zheng2022avatar} learns personalized implicit fields of blendshapes, pose correctives, and skinning weights, then formulates the avatar with linear summation of blendshapes followed by linear blend skinning. However, their linear formulation limits the amount of expression deformations.






\partitle{Geometry Anchored  Implicit 3D Representation}
Sparse local feature embedding attached on geometry has been demonstrated to be effective in improving the rendering quality of neural radiance field \cite{liu2020neural, liu2021neural, peng2021neural, zheng2022structured, lin2022fite}.  
It also naturally supports neural radiance field editing since the modification on the geometry can be directly propagated to the rendering \cite{NeuMesh}, which makes them a favorable representation to support the controllability for human avatar.
We adapt this representation to head avatar and incorporate head specific priors. Differently from prior art, we leverage a CNN in UV space to learn local, per-vertex features that are expression-dependent, improving generalization of out-of-train expressions.

\partitle{2D-based Head Avatars} There are numerous approaches that synthesize the head (or face) relying on 2D (explicit/implicit) representations, including 2D facial landmarks\cite{wang2018fewshotvid2vid, zakharov2019few, zakharov2020fast} and 2D warping fields \cite{wiles2018x2face, siarohin2019animating, siarohin2019first}. Landmark-based avatar models \cite{wang2018fewshotvid2vid, zakharov2019few, zakharov2020fast} synthesize the face conditioned to the facial landmarks extracted with a, usually pre-trained, landmark detector. Specifically, an encoder is applied to extract an identity embedding from a reference image, a decoder is adapted by the identity code to  animate the reference face with landmarks from the driving videos. X2Face \cite{wiles2018x2face} is the first approach to animate human heads by learning a dense warping field and producing the output video via image warping. MonkeyNet \cite{siarohin2019animating} and First Order Motion Model (FOMM) further propose to infer motion fields with self-learned keypoints, which significantly improves motion prediction and synthesizes higher quality renderings of heads. While most aforementioned methods can produce photorealistic results, they are not able to maintain geometry and multiview consistency due to their inherent 2D representation.

    

\section{Method}

Given a monocular RGB video containing $M$ frames $\{\bm{I}_1, \bm{I}_2, ..., \bm{I}_M\}$, our method reconstructs a head avatar representation that can be rendered under arbitrary facial expressions, head poses, and camera viewpoints.
We first pre-process the video to remove the background~\cite{pandey2021total,guo19} and obtain camera and 3DMM parameters for each frame. More specifically, we use FLAME~\cite{FLAME:SiggraphAsia2017} as the 3DMM and denote the fitted face with shape $\bm{\beta}$, expressions $\bm{\psi}_i$, poses $\bm{\theta}_i$ (\ie, neck, jaw, and eyes), where $i$ is the frame index, with which a head mesh can be obtained via $\bm{V}_i(\bm{\beta}, \bm{\psi}_i, \bm{\theta}_i)$.
Since $\bm{\beta}$ is fixed and does not depend on the pose or expressions for a given user, we omit it in the following sections for brevity.

An overview of our framework is shown in \cref{fig:overview}.
We adopt the 3DMM-anchored neural radiance field (NeRF) as the core representation for our head avatar (\cref{sec:avatar_representation}), where local features are attached to the vertices of the deformable 3DMM mesh.
During the inference, we first deform the 3DMM mesh based on the target configuration $\bm{V}_t=(\bm{\psi}_t, \bm{\theta}_t)$.
Then, for an arbitrary 3D query point, we aggregate the features from neighboring vertices on $\bm{V}_t$ to estimate the local density and color by Multi-Layer-Perceptrons (MLPs), which are then integrated in the volumetric rendering formulation to generate the color image.
To learn local features, we train CNN-based networks in the UV space to incorporate spatial context (\cref{sec:exp_dep_feats}). 
Our model is trained end-to-end with RGB supervisions (\cref{sec:training}).

\subsection{Avatar Representation}
\label{sec:avatar_representation}

\begin{figure}
    \centering
    \includegraphics[width=\linewidth]{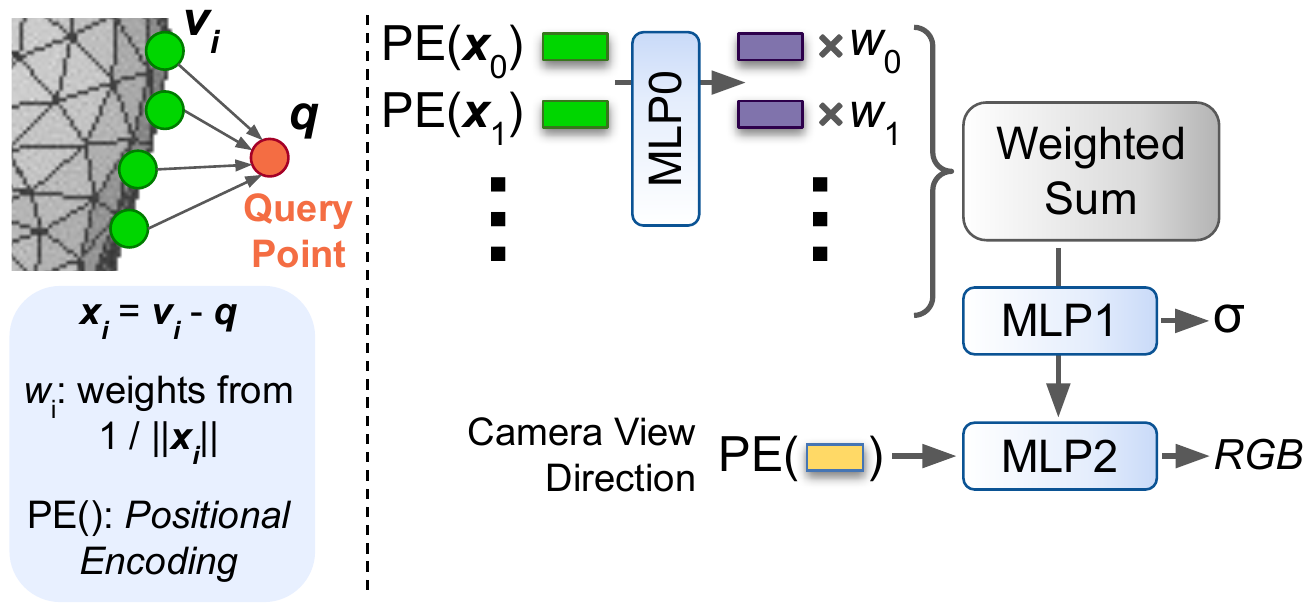}
    \vspace{-2em}
    \caption{Illustration of Avatar Representation (\cref{sec:avatar_representation}). Given a query point, we find its k-Nearest-Neighbor (k-NN) vertices from the 3DMM. Then, we decode these vertices and features into a density and color with respect to the input camera view direction, via Multi-Layer-Perceptrons (MLPs) interleaved with inverse-distance based weighted sum.
    }
    
    \label{fig:avatar_represen}
\end{figure}

An ideal representation for a head avatar should have the following properties: 1) Provides intuitive control to achieve the desired expression and head pose; 2) Requires a moderate amount of training data, \eg, a short monocular video; 3) Produces expression-dependent rendering details; 4) Generalizes reasonably well to unseen expressions. 

To this end, we propose the 3DMM-anchored neural radiance field (NeRF) as shown in \cref{fig:avatar_represen}. Inspired by local feature based neural radiance field \cite{yang2022neumesh,peng2021neural}, we attach feature vectors $\bm{z}^j$ on each 3DMM vertex $\bm{v}_i^j$ to encode the local radiance fields that can be decoded with MLPs, where $i$ denotes frame index and $j$ denotes vertex index.
In this way, the radiance field can be deformed according to vertex locations, hence can be intuitively controlled by the 3DMM expression and pose ($\bm{\psi}_i$, $\bm{\theta}_i$).
In addition, the 3DMM fitting on each frame
provides a rough tracking across deformable face geometries, such that all the frames can contribute into the learning of a unified set of local per-vertex features.
We will discuss model capacity and generalization in \cref{sec:exp_dep_feats}.



To decode the vertex features $\{\bm{z}^j\}$ into the radiance field for the frame $i$, given a 3D query point $\bm{q}$, we first find its $k$-Nearest-Neighbor ($k$-NN) vertices from the 3DMM mesh $\{\bm{v}_i^j\}_{j \in \mathcal{N}_k^{\bm{q}}}$ with attached features $\{\bm{z}^j\}_{j \in \mathcal{N}_k^{\bm{q}}}$. Then, we use two MLPs $\mathcal{F}_0$ and $\mathcal{F}_1$ with inverse-distance based weighted sum to decode local color and density. Formally,
\begin{align}
    \nonumber
    \hat{\bm{z}}_i^j &= \mathcal{F}_0 (\bm{v}_i^j - \bm{q}, \bm{z}^j) \\
    \hat{\bm{z}}_i &= \textstyle\sum_j w^j \hat{\bm{z}}_i^j \\
    \nonumber
    \bm{c}_i (\bm{q}, \bm{d}_i), \sigma_i (\bm{q}) &= \mathcal{F}_1 (\hat{\bm{z}}_i, \bm{d}_i),
\end{align}
where $w^j = \frac{d^j}{\sum_k d^k}$, $d^j = \frac{1}{ \| \bm{v}_i^j - \bm{q} \|_2}$ with $j \in \mathcal{N}_k^{\bm{q}}$, and $\bm{d}_i$ denotes the camera view direction. Finally, we render the output image with volumetric rendering formulation as in vanilla NeRF~\cite{mildenhall2021nerf} given the camera ray $\bm{r}(t) = \bm{o} + t \bm{d}$:
\begin{align}
    \bm{C}_i(\bm{r}) &= \int_{t_n}^{t_f} T(t) \sigma_i(\bm{r}(t)) \bm{c}_i(\bm{r}(t), \bm{d}) \diff t, \\
    \nonumber
    \text{ where } & T(t) = \text{exp} \left( - \int_{t_n}^{t} \sigma_i(\bm{r}(s)) \diff s \right)
\end{align}

To reduce misalignments caused by per-frame contents that cannot be captured by 3DMM (\eg, 3DMM fitting errors), we additionally learn error-correction warping fields during training inspired from prior works on deformable NeRF~\cite{jiang2022neuman,weng2022humannerf}. More specifically, we input the original query point and a per-frame latent code $\bm{e}_i$, which is randomly initialized and optimized during the training, into the error-correction MLPs $\mathcal{F}_{\mathcal{E}}$ to predict a rigid transformation, and apply it to the query point. The transformation is denoted as $\bm{q}^{\prime} = \mathcal{T}_i(\bm{q}) = \mathcal{F}_{\mathcal{E}}(\bm{q}, \bm{e}_i)$. Then we use the warped query point $\bm{q}^{\prime}$ to decode the color and density. Note that this warping field is disabled during testing. Please refer to the supplementary for detailed formulations of the warping field.

\begin{table*}[!t]
\small
\setlength{\tabcolsep}{0.6em}
\begin{tabular}{l|c|c|c|c|c}
\toprule
    & \textit{Subject0}     & \textit{Subject1}     & \textit{Subject2}     & \textit{Subject3}     & \textit{Subject4}     \\
    & \footnotesize LPIPS / SSIM / PSNR & \footnotesize LPIPS / SSIM / PSNR & \footnotesize LPIPS / SSIM / PSNR & \footnotesize LPIPS / SSIM / PSNR & \footnotesize LPIPS / SSIM / PSNR  \\
\midrule

TPSMM~\cite{zhao2022thin}       & 0.192 / 0.852 / 22.60 & 0.205 / 0.830 / \textbf{16.38} & 0.216 / 0.782 / 18.40 & 0.222 / 0.799 / 20.28 & 0.156 / 0.913 / 21.29 \\

FOMM~\cite{siarohin2019first}    & 0.171 / 0.841 / \textbf{22.93} & 0.179 / 0.827 / 16.02 & 0.202 / 0.777 / 18.98 & 0.186 / 0.798 / 22.28 & 0.122 / 0.915 / 23.94 \\

NHA~\cite{grassal2022neural}     & 0.165 / 0.836 / 20.20 & 0.166 / 0.840 / 15.48 & 0.178 / 0.809 / 17.99 & \textbf{0.153} / 0.798 / 21.31 & 0.091 / 0.926 / 23.78 \\

IMAvatar~\cite{zheng2022avatar}  & 0.207 / 0.852 / 21.26 & 0.187 / 0.848 / 15.98 & 0.265 / 0.729 / 15.80 & 0.214 / 0.782 / 20.37 & 0.142 / 0.897 / 20.63 \\

NerFACE~\cite{gafni2021dynamic} & 0.205 / 0.817 / 20.06 & 0.182 / 0.833 / 15.78 & 0.188 / 0.793 / 19.41 & 0.229 / 0.747 / 18.16 & 0.093 / 0.938 / 25.57 \\

Ours-D    & \textbf{0.144} / \textbf{0.864} / 21.92 & \textbf{0.152} / \textbf{0.855} / 16.23 & \textbf{0.141} / \textbf{0.841} / \textbf{20.42} & 0.156 / \textbf{0.833} / \textbf{23.05} & \textbf{0.075} / \textbf{0.944} / \textbf{25.71} \\
\bottomrule
\end{tabular}
\vspace{-0.5em}
\caption{Quantitative Comparison with state-of-the-art (SOTA) approaches. \textit{Subject4} is the data from NerFACE~\cite{gafni2021dynamic}, while other subjects are from our dataset. Our method achieves superior results than prior SOTAs.}
\label{tab:sota_quant}
\end{table*}

\subsection{Predicting Expression-Dependent Features}
\label{sec:exp_dep_feats}

While the proposed avatar representation (\cref{sec:avatar_representation}) enables intuitive controllability and convenience in learning, it still has  limited capability for modeling complex expression-dependent variations due to the use of frame-shared vertex features $\{\bm{z}^j\}$.


To overcome this, we propose to predict the dynamic vertex features $\{\bm{z}_i^j\}$ conditioned on the 3DMM expression and pose ($\bm{\psi}_i$, $\bm{\theta}_i$).
A common practice for NeRF-based methods is to use MLP-based architectures for dynamic feature prediction~\cite{gafni2021dynamic,zheng2022avatar,athar2022rignerf,zheng2022structured}. However, 
we find that this leads to blurry rendering results, presumably because of the limited model capacity due to the lack of spatial context (\ie, each vertex does not know the feature of its neighboring vertices).
Based on this intuition, we propose to process the 3DMM expression and pose ($\bm{\psi}_i$, $\bm{\theta}_i$) with CNNs in the texture atlas space (UV space) to provide local spatial context.

Specifically, we design two variations of CNN-based architecture to learn expression-dependent vertex features $\{\bm{z}_i^j\}$.
The first variant, denoted as \textit{Ours-C}, trains a decoder consisting of transposed convolutional blocks to predict the feature map in UV space directly from 1-D codes of 3DMM expression and pose.
We empirically find that such a model is effective in improving the overall rendering quality, however tends to fail and produce severe artifacts on out-of-training expressions (See discussions in \cref{sec:ablation}).
We then propose the second variant, denoted as \textit{Ours-D}, that uses the 3D deformation of 3DMM in UV space as the input for feature prediction, and observe that resulting avatar models are more resilient to stretchy and unseen expressions. 
Specifically, we first compute the vertex displacements using 3DMM expression and pose as $\bm{D}_i = \bm{V}_i(\bm{\psi}_i, \bm{\theta}_i) - \bm{V}_{neutral}(\bm{0}, \bm{0})$. We then rasterize the vertex displacements $\bm{D}_i$ into UV space, and process it with a U-Net $\mathcal{F}_{\mathcal{D}}$. Finally, the output UV feature map is sampled back to mesh vertices $\bm{V}_i$, serving as the dynamic vertex features $\{\bm{z}_i^j\}$ (\ie, the expression-dependent version of frame-shared vertex features $\{\bm{z}^j\}$ described in \cref{sec:avatar_representation}).

\if 
\subsection{Improving Mouth Region}
\label{sec:mouth_ext}

\begin{figure}[!t]
    \centering
    \includegraphics[width=\linewidth]{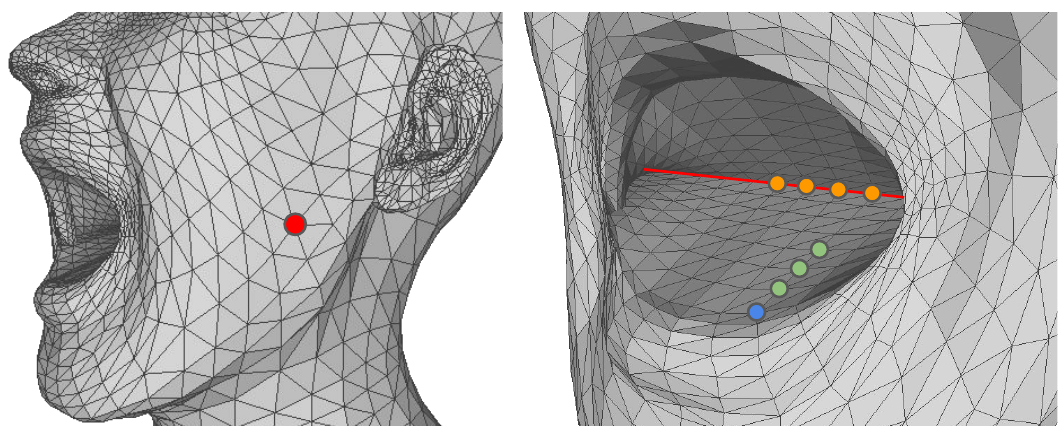}
    \caption{Illustration of interpolating the 3DMM mouth cavity.
    }
    \label{fig:mouth_ext}
\end{figure}

Due to the difficulty of collecting mouth interior data, most off-the-shelf 3DMMs do not cover the mouth interior, which degrades the mouth rendering quality. To this end, we propose a simple approach to extend the 3DMM mouth cavity with drivable geometry. As shown in \cref{fig:mouth_ext}, we first manually select $2$ anchor vertices (red point), one on each side of the the face to define a line (red line) where the innermost mouth vertices are positioned. We then linearly interpolate these vertices (orange points) on this line between the $2$ selected anchor vertices. We then interpolate the inner mouth vertices (green points) between the innermost vertices (orange points) and corresponding lip vertices (blue points). Lastly we mesh those points and integrate this inner mouth mesh with the rest of the face.
\fi

\subsection{Training Schema}
\label{sec:training}

Our model is trained on monocular RGB videos mainly with the photometric loss, where we penalize the $l_2$ distance between the rendering and the ground truth images. Formally $\mathcal{L}_{rgb} = \sum_i \sum_{\bm{r}} \| \bm{C}_i(\bm{r}) - \bm{I}_i(\bm{r}) \|_2$, where $\bm{r}$ denotes the camera ray of each pixel and $i$ denotes frame index.
To regularize the learning of error-correction warping field $\mathcal{T}(\bm{q})$, we adopt a elastic loss $\mathcal{L}_{elastic}$ similar to Nerfies~\cite{park2021nerfies}, and a magnitude loss defined as $\mathcal{L}_{mag} = \sum_{\bm{q}} \| \bm{q} - \mathcal{T}(\bm{q}) \|_2^2$. The total loss is defined as:
\begin{align}
    \mathcal{L} = \mathcal{L}_{rgb} + \lambda_{elastic} \mathcal{L}_{elastic} + \lambda_{mag} \mathcal{L}_{mag},
\end{align}
where we set $\lambda_{elastic} = 10^{-4}$ at the beginning and decay to $10^{-5}$ after $155$k iterations, and $\lambda_{mag} = 10^{-2}$. Please see supplementary for more details.

\if 
\subsection{Implementation Details}
\ft{Following [IMAvatar?], we apply [3D Tracking Method] to fit a FLAME face model for each frame. We further align the face model predicted from one training video by subtracting neck pose for each mesh vertex in order to mitigate the training difficulty. Like NeRF, our full model is hierarchical with the coarse and the fine network, which are simultaneously optimized by a photometric reconstruction loss.  To ensure stable training, we disable 3D warping field in the first 5k iterations, and to recover better fine-grained details, we adapt coarse-to-fine positional encoding in Nerfies. For optimization, we use the Adam optimizer with beta1, beta2. The batch size is set as ? and the learning rates is empirically set to. We train the model with total ? iterations for each identity.}

\label{sec:imp}
\begin{itemize}
    \item 256 UV size
    \item hierarchical as NeRF~\cite{} with corase and fine model
    \item Disable 3D warp field for first 5k iterations. Then use coarse to fine positional encoding as Nerfies.
    \item We discard the eyes poses from network inputs when predicting Expression-Dependent Features.
    \item We normalize the 3DMM vertices with its neck pose to align the head component in 3D space.
\end{itemize}
\fi

\section{Experiments}

\begin{figure*}[t]
    \centering
    \includegraphics[width=\linewidth]{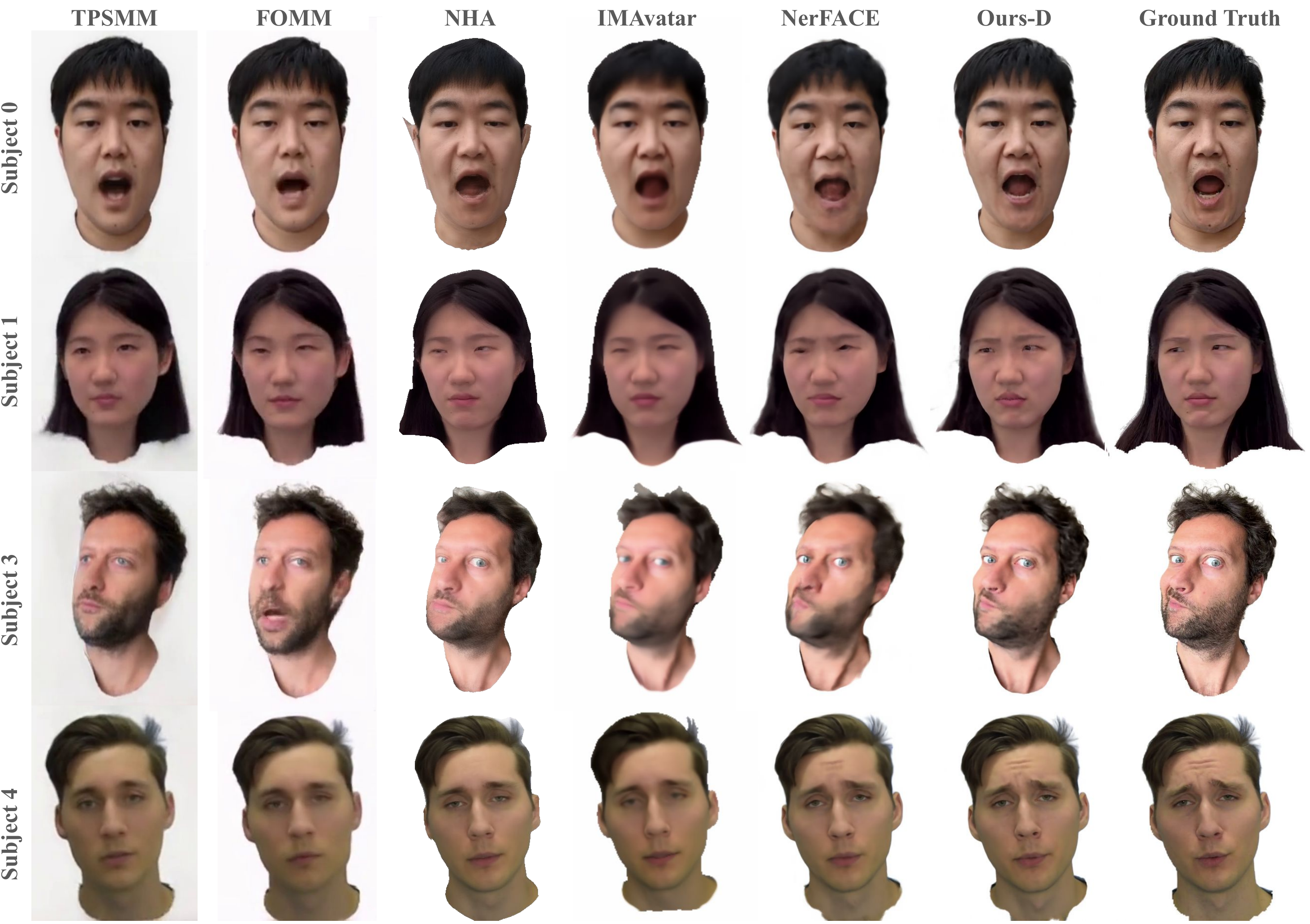}
    \vspace{-2em}
    \caption{Qualitative Comparison to prior state-of-the-art monocular head avatars. Note how our approach more faithfully reconstructs the ground truth expressions while preserving most of the high frequency details. 
    }
    
    \label{fig:vs_sota}
\end{figure*}

We train and evaluate our method on casually captured monocular RGB videos (\cref{sec:datasets_metrics}) and show that our method achieves superior rendering quality than prior state-of-the-art monocular RGB head avatars (\cref{sec:sota}). Then we verify our key observations on architectural choices to design a good avatar model, in terms of rendering quality and expression robustness (\cref{sec:ablation}).

\subsection{Datasets and Metrics}
\label{sec:datasets_metrics}

\partitle{Datasets}
Following the prior art~\cite{gafni2021dynamic}, we captured monocular RGB videos of various subjects with smartphones or webcams for training and evaluation. Each video is 1-2 minutes long (around 1.5k-2k frames at 30 FPS) with the first 1000-1500 frames as training data and the rest frames for evaluation if not otherwise specified. For the training clip, the subjects are asked to first keep a neutral expression and rotate their heads, then perform different expressions during the head rotation, with extreme expressions included. For the testing clip, the subjects are asked to perform freely without any constraints. To demonstrate that our method also works for common talking head videos, we also include a video from NerFACE~\cite{gafni2021dynamic}, which has significantly less variability in expressions when compared to our  capture protocol. We mask out background~\cite{pandey2021total} for each video, and obtain 3DMM and camera parameters with 3DMM fitting similar to that in NHA~\cite{grassal2022neural}. 
Please refer to the supplementary for examples of data. Please note that we collect relatively shorter training videos compared to related work. This favors a better user experience while still synthesizing high-quality avatars with more personalized and detailed expressions. 

\partitle{Metrics}
Following the prior art~\cite{gafni2021dynamic}, we use the following standard image quality metrics for quantitative evaluations: the Learned Perceptual Image Patch Similarity (LPIPS)~\cite{zhang2018unreasonable}, Structure Similarity Index (SSIM)~\cite{wang2004image}, and Peak Signal-to-Noise Ratio (PSNR).

\if 
Already have 3 subjects with our capture process (subject0-2) \url{https://drive.google.com/drive/folders/1hT8UOiw0qLD2l2Qb5mLMlN1Effd65h4v?resourcekey=0-U7nqigUHaC8k77uUz2l_Rw&usp=sharing}. Plan to add 1 from NerFACE (Subject~3) to show that our method works on common talking head. Capture 1-2 more and select 4 to 5 in total from them.

For each video, leave the last several frames for testing (depending on when testing clip starts during capture), and use the rest for training.

Metrics: PSNR, SSIM, LPIPS, FID.

\zb{2 camera capture then have GT on novel view synthesis.}

\zb{Visualize novel views results. fix one frame and move camera.}
\fi 

\subsection{Comparisons with State-of-Art}
\label{sec:sota}

We compare our method with five state-of-the-art methods of different types with publicly available implementations from the authors: NerFACE~\cite{gafni2021dynamic}, IMAvatar~\cite{zheng2022avatar}, NHA~\cite{grassal2022neural}, FOMM~\cite{siarohin2019first}, and TPSMM~\cite{zhao2022thin}. For subject-specific methods (\ie Ours, NerFACE~\cite{gafni2021dynamic}, IMAvatar~\cite{zheng2022avatar}, and NHA~\cite{grassal2022neural}), we train the avatar for each subject separately with training frames of each video, then drive and render the trained avatar with 3DMM and camera parameters of testing frames.
For few-shot methods (\ie, FOMM~\cite{siarohin2019first}, and TPSMM~\cite{zhao2022thin}), we use the first frame of each video, which shows a frontal head, as the source image, then use testing frames as driving images in self-reenactment manner. Finally, the generated images of each method are compared with ground truth testing frames quantitatively (See \cref{tab:sota_quant}) and qualitatively (See \cref{fig:vs_sota}).

As shown in \cref{fig:vs_sota},
FOMM~\cite{siarohin2019first} and TPSMM~\cite{zhao2022thin} struggle with large head rotations and fail to produce 3D consistent results.
NHA~\cite{grassal2022neural} uses explicit mesh surface with neural textures. The fix mesh topology makes it hard to handle challenging hair in \textit{Subject~3}. Also, NHA uses linear 3DMM blendshapes, making it hard to capture complex expressions (\eg, \textit{Subject~1 \& 3}) and wrinkles (\eg, \textit{Subject~4}).
IMAvatar~\cite{zheng2022avatar} uses an occupancy field to model geometry, making it hard to handle challenging hair. Despite learning a personalized blendshape field, the linear formulation still limits their model capacity of handling complex expressions. In addition, we observed training instability of IMAvatar on the captured data, converging to oversmooth results.
NerFACE~\cite{gafni2021dynamic} works relatively well on data with easy expressions (\ie, common talking heads in \textit{Subject~4}), but struggles on our challenging data and estimates incorrect geometry (\eg, distorted head for \textit{Subject~3}) and blurry rendering.
In contrast, our method gives superior results on all the aspects discussed above. \cref{tab:sota_quant} further quantitatively confirms the good rendering quality of our method.



\subsection{Driving the Avatar}
\label{sec:drive}

\begin{figure}[t]
    \centering
    \includegraphics[width=\linewidth]{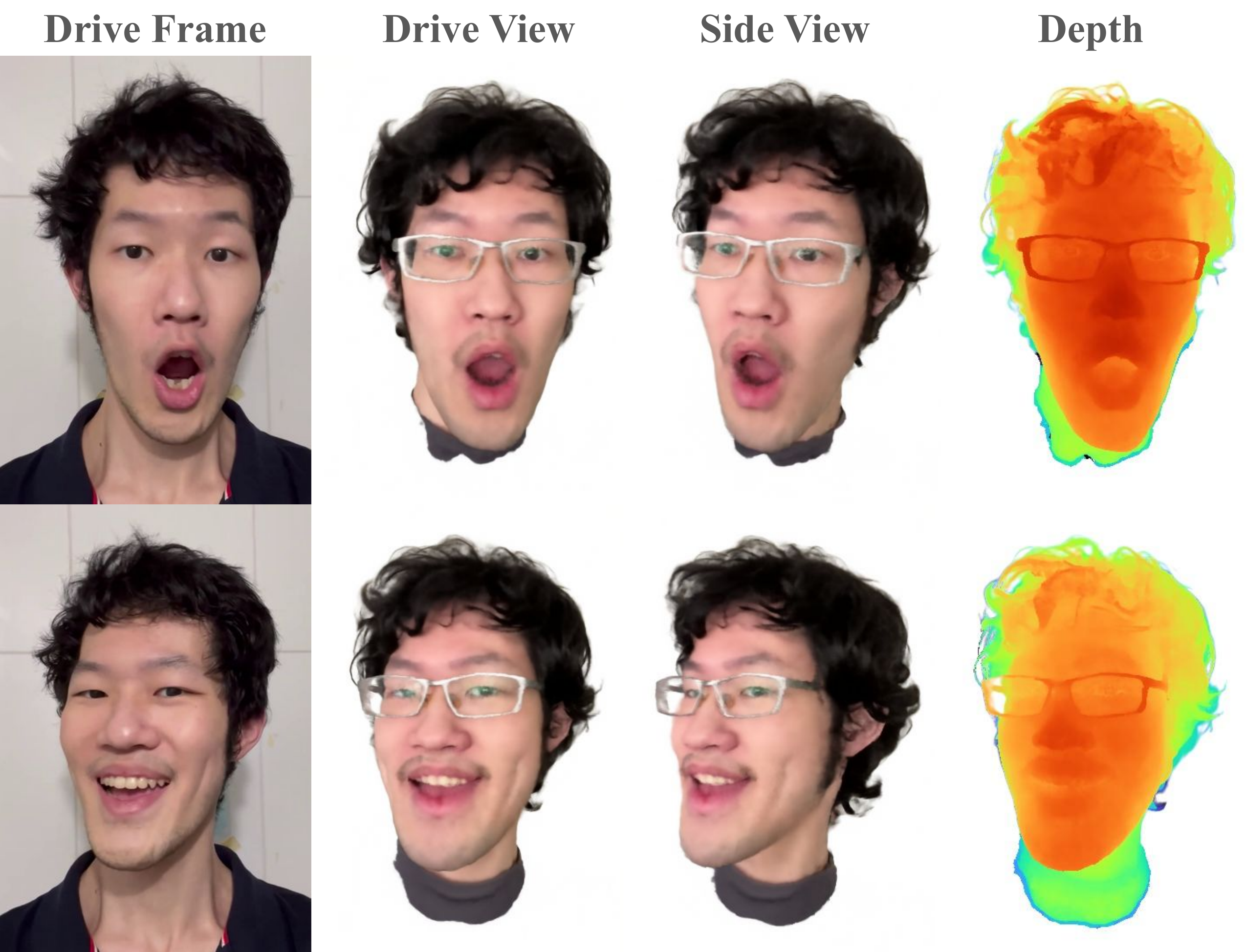}
    \vspace{-2em}
    \caption{Results on driving the learned avatar by the same subject under different capture conditions. Our method produces faithful expressions, multi-view consistent rendering, and good geometry.
    }
    
    \label{fig:drive}
\end{figure}

\begin{figure}[t]
    \centering
    \includegraphics[width=\linewidth]{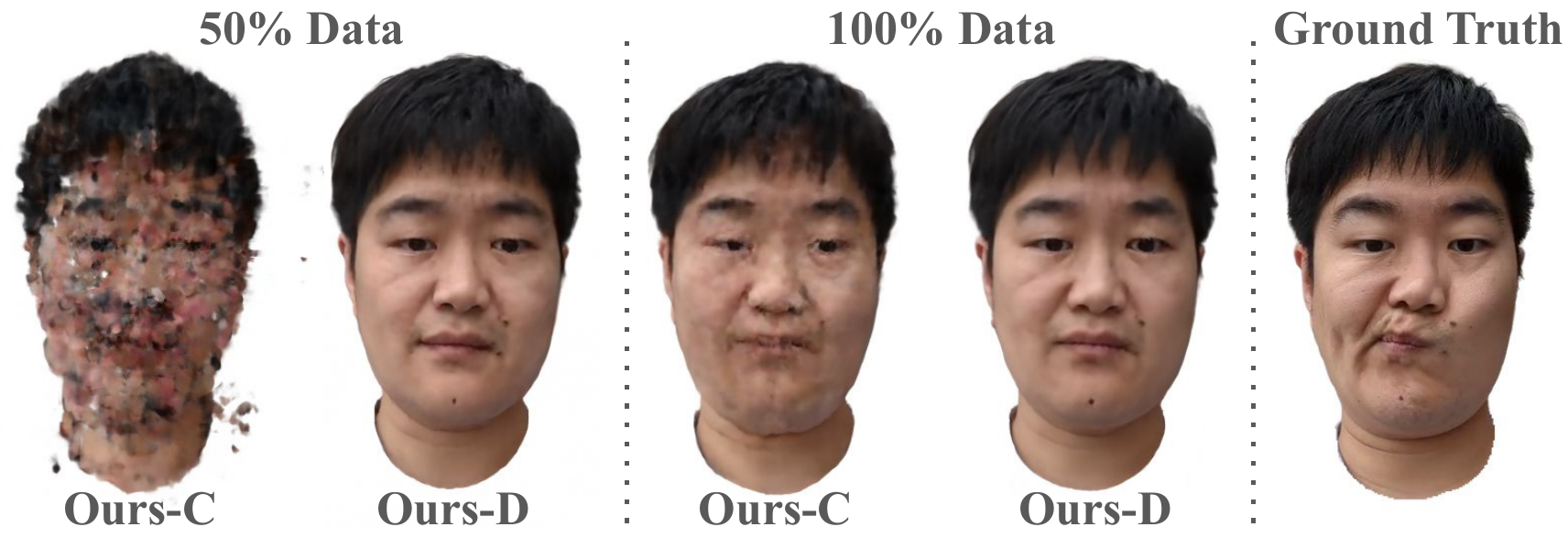}
    \vspace{-2em}
    \caption{Results on expressions out of the training distribution with different amount of training data. \textit{Ours-D} more robustly handles unseen expressions and degrades less with fewer training data.
    }
    \label{fig:ablation_exp_robust}
\end{figure}

After training, the learned avatar model can be driven by the same subject under different capture conditions, \eg hair style, illumination, glasses. We show the driving results in \cref{fig:drive}. Our avatar faithfully reproduces the expressions of the driving frame, while also achieving multi-view consistent renderings and generating high quality geometry.

\subsection{Ablation Study: Expression Features}
\label{sec:ablation}

\begin{figure*}[t]
    \centering
    \includegraphics[width=\linewidth]{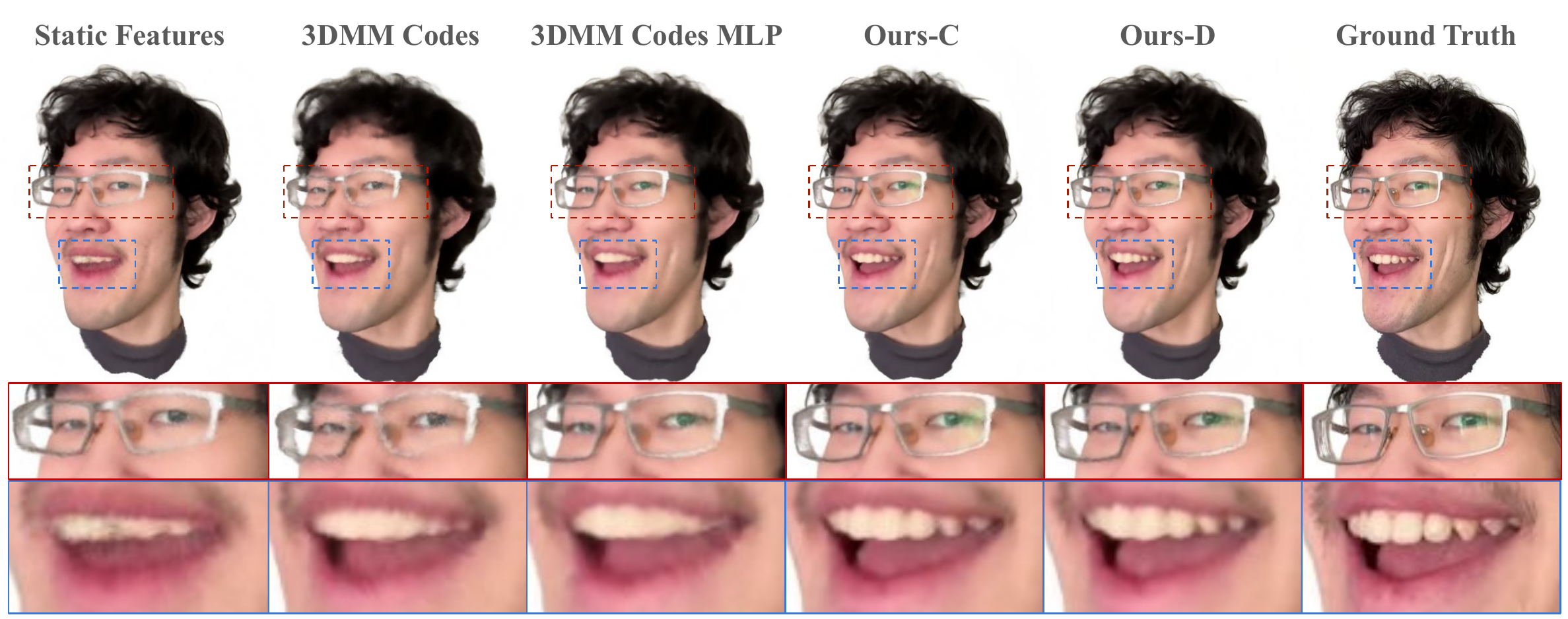}
    \vspace{-2em}
    \caption{Comparison between different designs for local vertex feature learning. See \cref{sec:ablation} for more details. ``Static feature'' struggles on capturing personalized expressions. ``3DMM Codes'' improves the personalization but suffers from overall blurriness. ``3DMM Codes MLP'' further improves the sharpness, but still cannot present the details. Overall, our convolution-based methods lead to superior renderings on areas such as cheek silhouette, glasses frames and reflections, and teeth.
    }
    \label{fig:ablation}
\end{figure*}

Learning good local features is crucial for improving model capacity and capturing high frequency details, without losing the regularization from 3DMM (\cref{sec:exp_dep_feats}).
We investigate and compare three alternative approaches and our two model variations for feature learning:

\partitle{Static Features}
We use frame-shared (hence expression-shared) vertex features $\{\bm{z}^j\}$.
As shown in \cref{fig:ablation}, the expression-shared features struggle to capture strong expression-dependent variations, leading to incorrect geometry (\eg, incorrect cheek silhouette), blurry details (\eg, teeth), and inferior LPIPS scores in \cref{tab:ablations}.

\partitle{3DMM Codes}
We extend \textit{Static Features} by directly concatenating 3DMM expression and pose codes ($\bm{\psi}_i$, $\bm{\theta}_i$) to the vertex features $\{\bm{z}^j\}$. 
This allows the model to have more capacity to capture expression-dependent variations. 
Although median-level expression characteristics are recognizable (\eg, cheek silhouette in \cref{fig:ablation}), the results are even blurrier for components less depend on expressions such as hairs, possibly due to the limited model capacity of shallow MLPs without spatial context.

\partitle{3DMM Codes MLP}
To further increase the model capacity, we use a more sophisticated MLP architecture inspired from Zheng et al.~\cite{zheng2022structured}, which is a MLP-based conditional Variational AutoEncoder (cVAE), to predict the expression-dependent vertex features $\{\bm{z}_i^j\}$ from the 3DMM expression and pose codes ($\bm{\psi}_i$, $\bm{\theta}_i$).
More specifically, the cVAE is conditioned on 3DMM codes and vertex coordinates on neutral face mesh $\bm{V}_{neutral}$. Then, the cVAE encodes the frame index into a latent code and decodes it into the vertex features $\{\bm{z}_i^j\}$. Although the overall sharpness of results is improved (\eg, glasses and hair in \cref{fig:ablation}), this model still produces blurry details such as teeth, eyes, and wrinkles. \cref{tab:ablations} further confirms that this more sophisticated MLP architecture still produces  inferior results compared to CNN-based methods \textit{Ours-C} and \textit{Ours-D}.

\begin{table}[t]
\label{tab:ablations}
\centering
\small
\setlength{\tabcolsep}{0.4em}
\begin{tabular}{l|c|c|c|c}
\toprule
                              & \textit{Subject0} & \textit{Subject1} & \textit{Subject2} & \textit{Subject3} \\
\midrule
\multicolumn{5}{c}{\footnotesize \textit{Full Training Data}} \\
\hline

Static Features             & 0.1559            & 0.1586            & 0.1552            & 0.1688            \\

3DMM Codes              & 0.1599            & 0.1620            & 0.1746            & 0.1738            \\

3DMM Codes MLP  & 0.1568            & 0.1551            & 0.1505            & 0.1686            \\

Ours-C           & \textbf{0.1417}            & \textbf{0.1457}            & \textbf{0.1383}            & \textbf{0.1550}            \\

Ours-D                          & \underline{0.1439}            & \underline{0.1523}            & \underline{0.1415}            & \underline{0.1559}            \\

\hline
\multicolumn{5}{c}{\footnotesize \textit{First $50\%$ Training Data}} \\
\hline

Ours-C & 0.2038            & 0.1483            & 0.1580            & 0.1566           \\
& \textcolor{red}{+0.0621} & \textcolor{red}{+0.0026} & \textcolor{red}{+0.0197} & \textcolor{red}{+0.0016} \\

Ours-D                & 0.1711            & 0.1511            & 0.1516            & 0.1558            \\
& \textcolor{red}{+0.0272} & \textcolor{Green}{-0.0012} & \textcolor{red}{+0.0101} & \textcolor{Green}{-0.0001} \\

\bottomrule

\end{tabular}
\caption{Quantitative ablations with baselines in LPIPS scores (lower is better). \textbf{Bold} denotes the best while \underline{underline} denotes the second best.}
\label{tab:ablations}
\end{table}

\partitle{Ours-C \textit{vs.} Ours-D}
By replacing MLPs with CNNs in UV space (Please refer to \cref{sec:exp_dep_feats} for more details), both variations of \textit{Ours} achieve superior rendering quality than MLP-based baselines, as shown in \cref{fig:ablation}.
Moreover, we find that \textit{Ours-D} is more robust to expressions that are outside the training distribution compared to \textit{Ours-C} (See \cref{fig:ablation_exp_robust}).
To further investigate the model robustness to out-of-training expressions, we train \textit{Ours-C} and \textit{Ours-D} with the first $50$\% of training frames, to simulate the scenario with less expression coverage during training. As shown in \cref{tab:ablations} and \cref{fig:ablation_exp_robust}, \textit{Ours-D} degrades less than \textit{Ours-C}, indicating enhanced robustness.


\subsection{Robustness to Expression Extrapolation}
\label{sec:exp_extrap}

\begin{figure}[t]
    \centering
    \includegraphics[width=\linewidth]{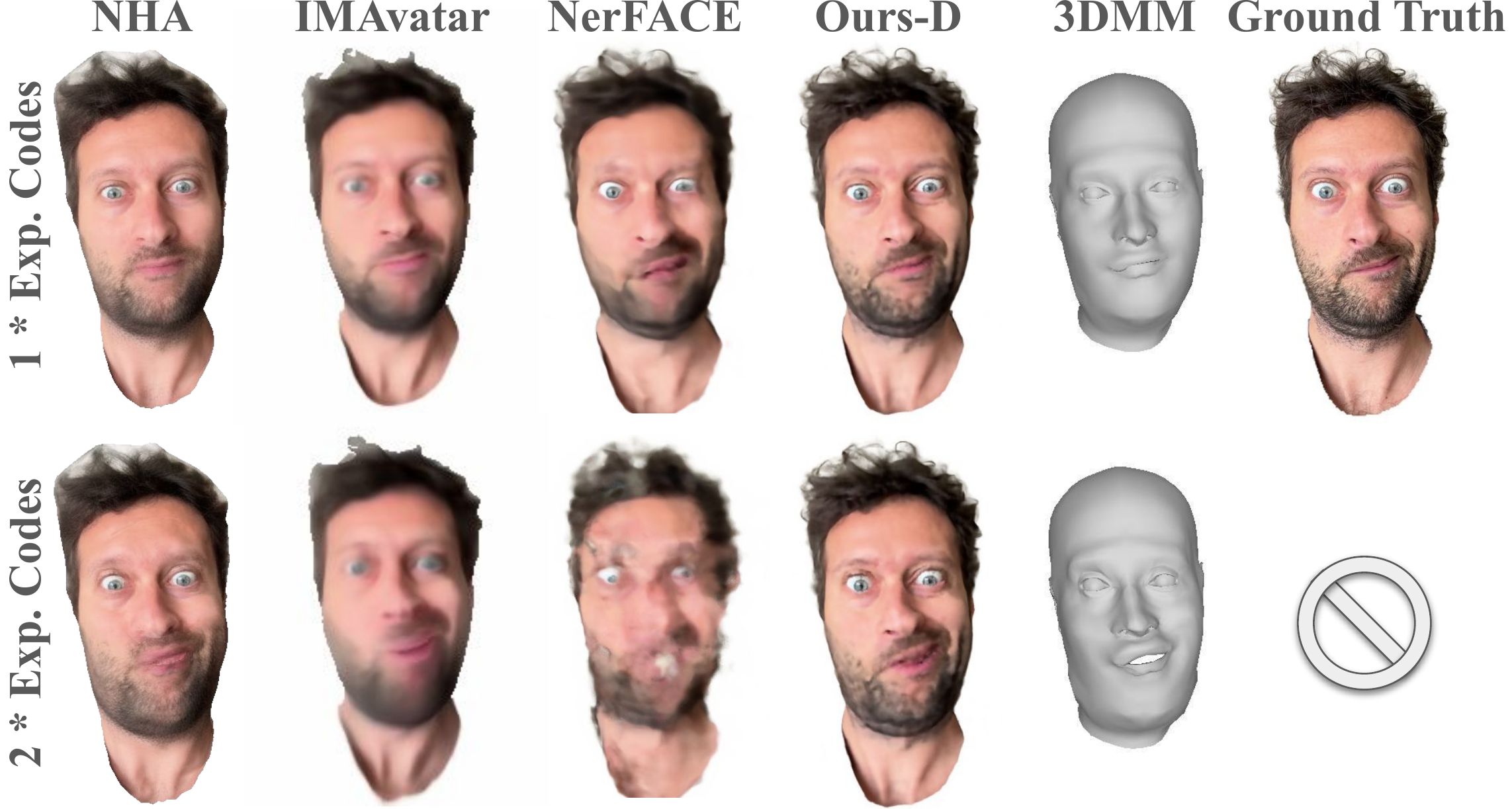}
    \caption{Expression extrapolation results and comparisons. Compared to other work, our method performs well on extreme and out-of-training expression.
    }
    \label{fig:exp_extrap}
\end{figure}

We qualitatively test our method on expression extrapolation setting by artificially manipulating the expression code for a given test frame. In particular, we double the value of the expression code and compare the results with prior works. As shown in \cref{fig:exp_extrap}, NHA~\cite{grassal2022neural} gives reasonable expression extrapolation results for both geometry and appearance. However, it cannot faithfully capture the out-of-3DMM details when compared to the ground truth. IMAvatar~\cite{zheng2022avatar} gives reasonable geometry for extrapolation, but struggles to produce sharp extrapolated appearances. NerFACE~\cite{gafni2021dynamic} fails in producing reasonable renderings. In contrast, our method not only faithfully captures out-of-3DMM details, but also generalizes to extrapolated expressions.


    

\if
\partitle{w/ vs. w/o unet}

\partitle{vertex 3D deformations of expressions and poses + Unet in UV space vs. 3DMM expression and pose parameters + MLPs}

\partitle{w/ vs. w/o mouth cavity interpolation on 3DMM}

\partitle{knn vs. project query point to mesh surface}
\fi

\if
\subsection{Some issues on experiments}
(1) 3DMM fitting on testing frames are not perfect, thus having misalignment between predictions and GT.

\zb{get numbers first. Then see whether we need to get better 3DMM fitting with training on all frames.}

\section{last week introduction}
Animatable human avatar is one of the fundamental techniques for many downstream applications, such as AR/VR communication, Virtual try on, game and movie industry.

While obtaining high quality avatar is already possible, existing solutions usually require bulky equipment setup such as camera arrays and light stages, or involve tedious manual intervention.

The recent integration of parametric model and volumetric representation is closing the quality gap between light-weight captured avatar and high-end productions.

One typical approach is attaching learnable latent codes on the mesh vertices of SMPL body model, then decode the code-attached SMPL into radiance fields and perform volumetric rendering to obtain final images.
This design introduces temporal constraints via sharing the same codes across different body poses, and also leverages SMPL meshes as a shape prior.
Following works further input SMPL pose parameters to capture pose-dependent details.

Given the encouraging results it gives on body avatar, further adapting this “code-attached mesh + volumetric decoding” design to monocular head avatar can be a promising direction. Although directly adapting this design to head avatar can give reasonable results, we argue that it does not fully leverage the strong prior of human head.

To this end, we propose our method.

We use different constraints to regularize the ill-posedness of monocular reconstruction, while allowing the flexibility to capture expression-dependent details.

Direct adapting: Expression-shared codes thus no expression dependent details; or 3DMM expression parameters + MLPs that is too abstract and do not contain spatial information.

Ours: vertex 3D deformations of expressions + Unet in UV space.
An interpolation method to create coarse mouth cavity on 3DMM to better capture mouth interior.
\fi

\section{Discussion}
In this work, we presented a framework to learn high-quality controllable 3D head avatars from monocular RGB videos. The core of our method is a 3DMM-anchored neural radiance field decoded from features attached on 3DMM vertices. The vertex features can be either predicted from 3DMM expression and pose codes, or vertex displacements, via CNNs in UV space, where the former favors quality and the latter favors robustness. We experimentally demonstrate that it is possible to learn high-quality avatars with faithful, highly non-rigid expression deformations purely from monocular RGB videos, leading to a superior rendering quality of our method compared to other approaches.

\paragraph{Limitations.}
Compared to the state-of-the-art approaches, our proposed framework learns portrait avatars with superior rendering quality. Nevertheless, our method still inherits the disadvantages of NeRF~\cite{mildenhall2021nerf} on time-consuming subject-specific training and slow rendering. Our method relies on the expression space of a 3DMM, thus cannot capture components that are completely missing in the 3DMM, such as the tongue. Moreover, extending our method to include the upper body or integrating into full body avatars are also interesting future directions.

\paragraph{Ethical Considerations.}
The rapid progress on avatar synthesis facilitates numerous downstream applications, but also raises concerns on ethical misuses. On the synthesis side, it would be ideal to actively use watermarking and avoid driving avatars with different identities. On the viewing side, forgery detection~\cite{tolosana2020deepfakes,agarwal2020detecting,rossler2018faceforensics,rossler2019faceforensics++,li2020celeb} is an active research field with promising progress. However, it is still hard to guarantee reliable detection of fake visual materials at the current stage. Encouraging the use of cryptographical signatures may also be a potential solution to ensure the authenticity of visual material.

{\small
\bibliographystyle{ieee_fullname}
\bibliography{egbib}
}

\clearpage

\begin{center}
\vspace{10mm}
\textbf{\Large Supplementary Materials}
\end{center}
\renewcommand\thesection{\Alph{section}}
\renewcommand\thetable{\Alph{table}}
\renewcommand\thefigure{\Alph{figure}}

\setcounter{section}{0}
\setcounter{figure}{0}
\setcounter{table}{0}

\begin{figure*}[t]
    \centering
    \includegraphics[width=\linewidth]{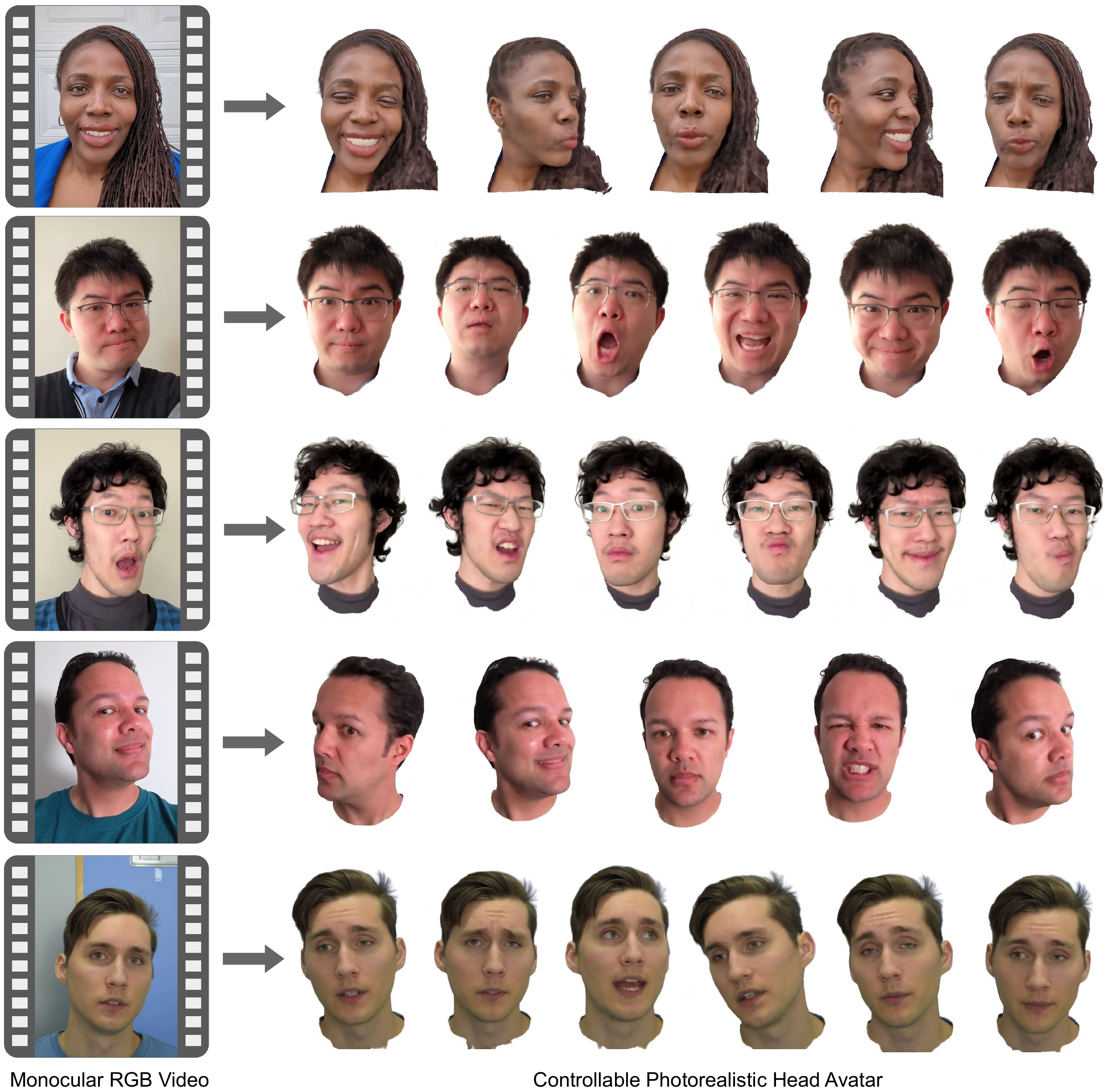}
    \caption{We propose a method to build a 3D avatar representation of a person using just a single short monocular RGB video (e.g., 1-2 minutes), which can be rendered with user-defined expression and viewpoint. Note how our method captures extreme expressions and fine scale facial details. Please check our supplementary webpage for more video results, and discussions on the limitation.
    }
    \label{fig:ours_supp}
\end{figure*}

We provide additional information in this supplementary material, including Warp Field Formulation (\cref{sec:warp_field}), Implementation Details (\cref{sec:impl}), Examples of Data (\cref{sec:data_example}), as well as Image and Video Results (\cref{fig:ours_supp}, \cref{sec:more_results}, and the accompanying supplementary webpage). Please see our project webpage \href{https://augmentedperception.github.io/monoavatar/}{augmentedperception.github.io/monoavatar} for more results.

\section{Warp Field Formulation}
\label{sec:warp_field}

\partitle{Motivation}
Though the 3DMM fitting can reasonably track the head and expression motions, there are still ad-hoc motions that cannot be handled by the 3DMM, such as the hair movements and tracking errors, which lead to misalignments between the 3DMM mesh and images and cause the model to learn blurred appearances.

As described in Sec. 3.1, inspired from prior works on deformable NeRF~\cite{jiang2022neuman,weng2022humannerf}, we learn error-correction warp fields with small magnitudes during training to reduce the misalignments, enabling the model to learn sharper appearances. During testing, we discard the warp fields since they are overfit to training frames. Since the warp fields are small in magnitudes (encouraged by the loss function $\mathcal{L}_{mag}$ in Eq.3), they do not affect the inference heavily. As a result, the renderings are equally sharp, albeit with slightly miss-aligned finer details compared to the ground truth.


\partitle{Formulation}
We input the original query point $\bm{q}$ and a learnable per-frame latent code $\bm{e}_i$ ($i$ is frame index) into the error-correction MLPs $\mathcal{F}_{\mathcal{E}}$ to predict a rigid transformation. The rigid transformation contains a rotation $\bm{R} \in SO(3)$ , a rotation center $\bm{c}^{rot}$, and a translation $\bm{t}$. Finally, the rigid transformation is applied to the query point to obtain the warped point $\bm{q}^{\prime}$. Formally, we have

\begin{align}
    \bm{R}, \bm{c}^{rot}, \bm{t} &= \mathcal{F}_{\mathcal{E}} \left( \bm{q}, \bm{e}_i \right) \\
    \bm{q}^{\prime} &= \bm{R} \left( \bm{q} + \bm{c}^{rot} \right) - \bm{c}^{rot} + \bm{t},
\end{align}
where $\bm{R}$ is parameterized by a pure log-quaternion predicted by the MLPs. We denote the full transformation as $\bm{q}^{\prime} = \mathcal{T}_i(\bm{q}) = \mathcal{F}_{\mathcal{E}}(\bm{q}, \bm{e}_i)$. Then, the warped point is used as the query point to decode the density and color as described in Sec. 3.1. Note that this warping field is only used during training and disabled during testing.


\section{Implementation Details}
\label{sec:impl}

To improve the training convergence, we remove the background \cite{pandey2021total,Guo2019The} and align the head in 3D space by normalizing the 3DMM vertices with its neck pose. Similar to NeRF \cite{mildenhall2021nerf}, our full model is hierarchical with the coarse and the fine networks, which are simultaneously optimized by a photometric reconstruction loss. To ensure stable training, we disable 3D warping field in the first 5k iterations, and 
enable it in the following iterations. For optimization, we use the Adam optimizer with $\beta_{1}=0.9$, $\beta_{2}=0.999$. The batch size is set as $1024$ rays and the learning rates are empirically set to: (1) $10^{-4}$ and exponentially decay to $10^{-5}$ after $400$k for warp field networks. (2) $10^{-3}$ and exponentially decay to $10^{-4}$ after $400$k for other networks. We train the model with total $400$k iterations for each subject. We adapt coarse-to-fine positional encoding (as used in Nerfies~\cite{park2021nerfies}) on the coordinate input of the warp field networks for better training stability. More specifically, we start with $0$ frequency bands and linearly increase to $6$ after $80$k iterations. For other modules, we adapt positional encoding as in NeRF~\cite{mildenhall2021nerf} with $10$ frequency bands on all coordinate inputs and $4$ on camera views.

\subsection{Network Architecture}
As detailed in the main paper, the framework consists of three modules: a 3DMM-anchored NeRF, a expression-dependent feature predictor, and a warping field predictor.

\subsubsection{3DMM-anchored NeRF}

\begin{figure}[th]
    \centering
    \includegraphics[width=\linewidth]{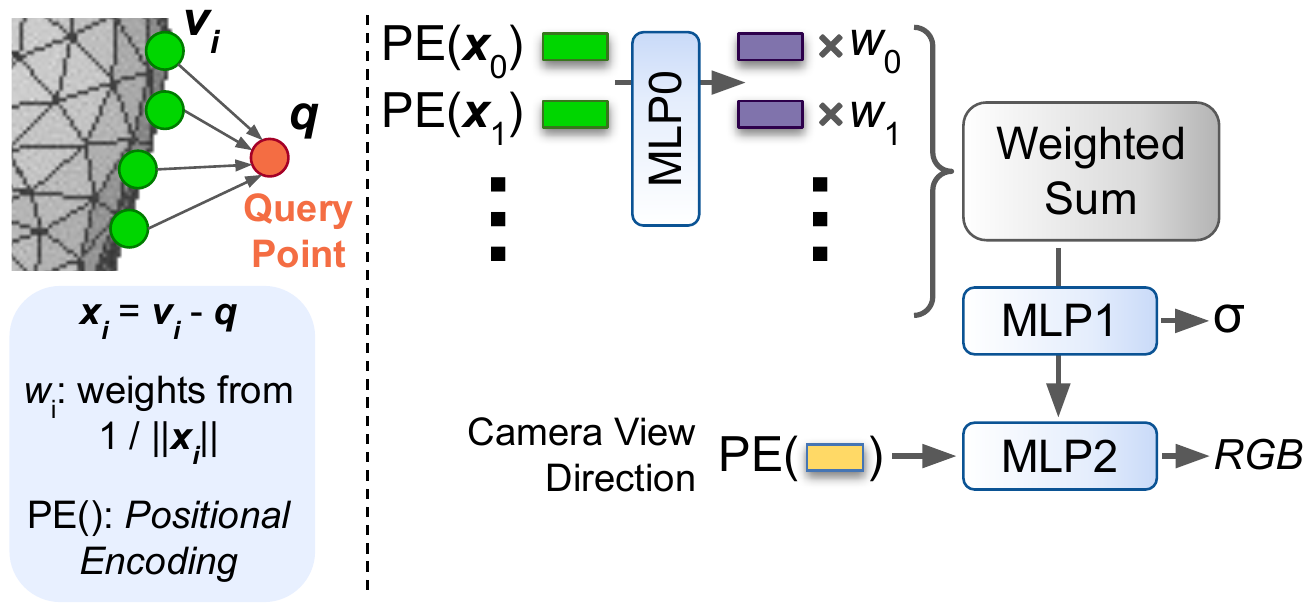}
    \caption{Illustration of Avatar Representation. Given a query point, we find its k-Nearest-Neighbor (k-NN) vertices from the 3DMM. Then, we decode these vertices and features into a density and color with respect to the input camera view direction, via Multi-Layer-Perceptrons (MLPs) interleaved with inverse-distance based weighted sum.
    }
    
    \label{fig:avatar_represen}
\end{figure}

As described in Sec. 3.1 of the main paper, we adopt the 3DMM-anchored neural radiance field (NeRF) to represent our head avatar. As shown in \cref{fig:avatar_represen}, we attach 64-dimensional feature vectors on each vertex of the FLAME model \cite{FLAME:SiggraphAsia2017}, which are predicted from the U-Net described in Sec. 3.2. During inference, we first concatenate the normalized coordinates $\bm{v}_i^j - \bm{q}$ (positional encoded) of the vertex with it's corresponding attached features and pass them into the MLP0, which comprises 3 hidden layers with 128 neurons each and applies ReLU activation, to produce latent features. We then aggregate the latent features of the nearest 4 vertices by a inverse-distance based weighted sum. The aggregated feature is then decoded into density and color with 2 branches. For density, the aggregated feature is decoded by MLP1 + a Fully Connected (FC) layer. For color, the aggregated feature is decoded by MLP1 + MLP2. MLP1 comprises 3 hidden layers with 128 neurons each and applies ReLU activation. MLP2 comprises 1 hidden layers with 64 neurons and 1 FC layer with 3 outputs. To handle view-dependent effects, we also pass the ray view direction (positional encoded) into the MLP2 to decode the RGB color.

\subsubsection{Expression-Dependent Features Predictor}
Our expression-dependent features predictor is a 6-level residual U-Net. We use residual blocks to extract feature, and the feature channels of each level are set as 8, 16, 32, 64, 128, 256. In the decoder, residual blocks with transposed convolutions are applied to increase the spatial resolution. The leaky ReLU is applied after each convolutional layer with slope $0.2$. The input of the predictor is a 3D deformation map in $256 \times 256$ resolution which stores the vertex displacements from the neural expression to the current facial expression in UV space.

\subsubsection{Warping Field Predictor}
The error-correction MLPs $\mathcal{F}_{\mathcal{E}}$ is utilized to predict error-correction warping fields to reduce misalignments from 3DMM and improve the training. It consists of 5 hidden layers with 128 neurons each, followed by ReLU activation, then 3 branches of two-layers MLPs with 128 neurons are added at the end for regressing each output (as described in \cref{sec:warp_field}: pure log-quaternion of the rotation (\ie, SO(3)) $\bm{R}$, rotation center $\bm{c}^{rot}$, and translation $\bm{t}$).

\subsection{3DMM Fitting Details}
We have implemented the same optimization-based fitting algorithm as NHA~\cite{grassal2022neural} with the following differences: We 1) used MediaPipe for improved nose, eyes, and eyebrows landmarks;
2) re-initialized camera poses (by Perspective-n-Point) and expression parameters (to neutral) every $200$ frames to prevent local optima;
3) increased optimization steps per frame to accommodate for more challenging expressions in our data.
Note that we use the same fitting results across all methods for a fair comparison.

\subsection{Video Capture Protocol}
\begin{figure}[t]
    \centering
    \includegraphics[width=\linewidth]{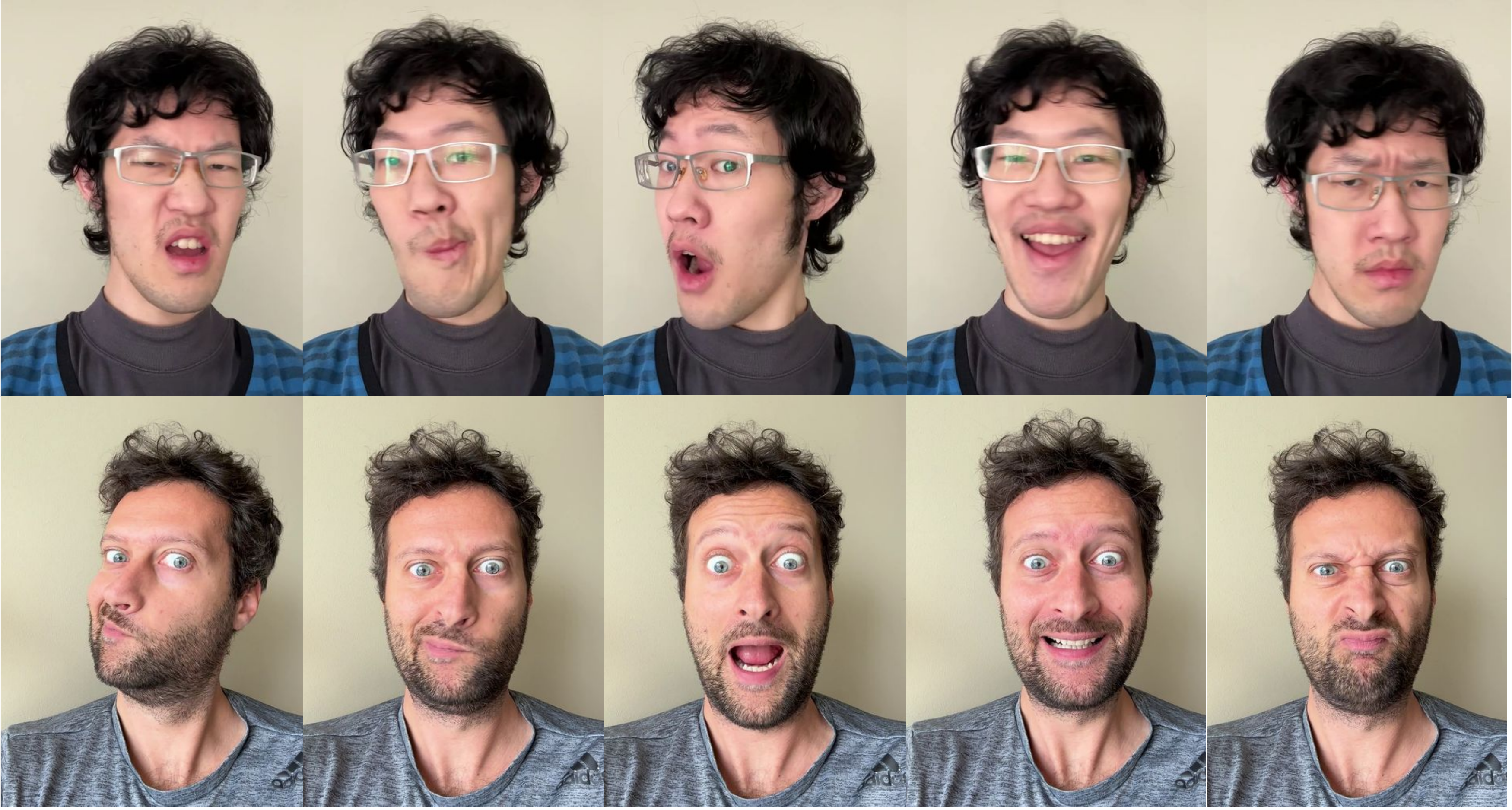}
    \caption{Examples of reference expressions for video capture.
    }
    
    \label{fig:ref_exps}
\end{figure}

We ask users to capture 1-2 min selfie videos with high resolution (over $500 \times 500$ pixels in the head) under well-lit conditions using phone/webcam, following instructions below (the same as in Sec.4.1). For the training clip, the users are asked to first keep a neutral expression and rotate their heads, then perform different expressions during the head rotation, with extreme expressions included. For the testing clip, the users are asked to perform freely without any constraints. We provide several reference expressions shown in \cref{fig:ref_exps} for users to follow, but users are not asked to strictly perform the same expressions.

\if
\label{sec:imp}
\begin{itemize}
    \item 256 UV size
    \item hierarchical as NeRF~\cite{} with coarse and fine model. 256 coarse 128 fine.
    \item Disable 3D warp field for first 5k iterations. Then use coarse to fine positional encoding as Nerfies.
    \item We discard the eyes poses from network inputs when predicting Expression-Dependent Features.
    \item We normalize the 3DMM vertices with its neck pose to align the head component in 3D space.
    \item first 500 iter use L2 loss. Then L2-norm loss.
    \item UNet: resblock: conv + lrelu, 256x256 in down 6 blocks, 4x4 bottleneck, up 6 blocks, 256x256 out. Down 8, x2, to 256 then all 256. Up 256, /2, to 64 then all 64.
    \item MLPs: Before weighted sum MLP 3 layers 128 channel. After weighted sum MLP 3 layers 128 channel. density 1 FC. RGB feat\_ch + view direction to 64 to 3. All MLP use relu.
    \item verts\_feats 64 channel. 
    \item batch number of rays 1024 from 8 images.
    \item warp: latent code 64. 5 shared layers . Then 2 layers for each output. All 128 ch. relu.
\end{itemize}
\fi

\section{Examples of Data}
\label{sec:data_example}

\begin{figure*}[t]
    \centering
    \includegraphics[width=\linewidth]{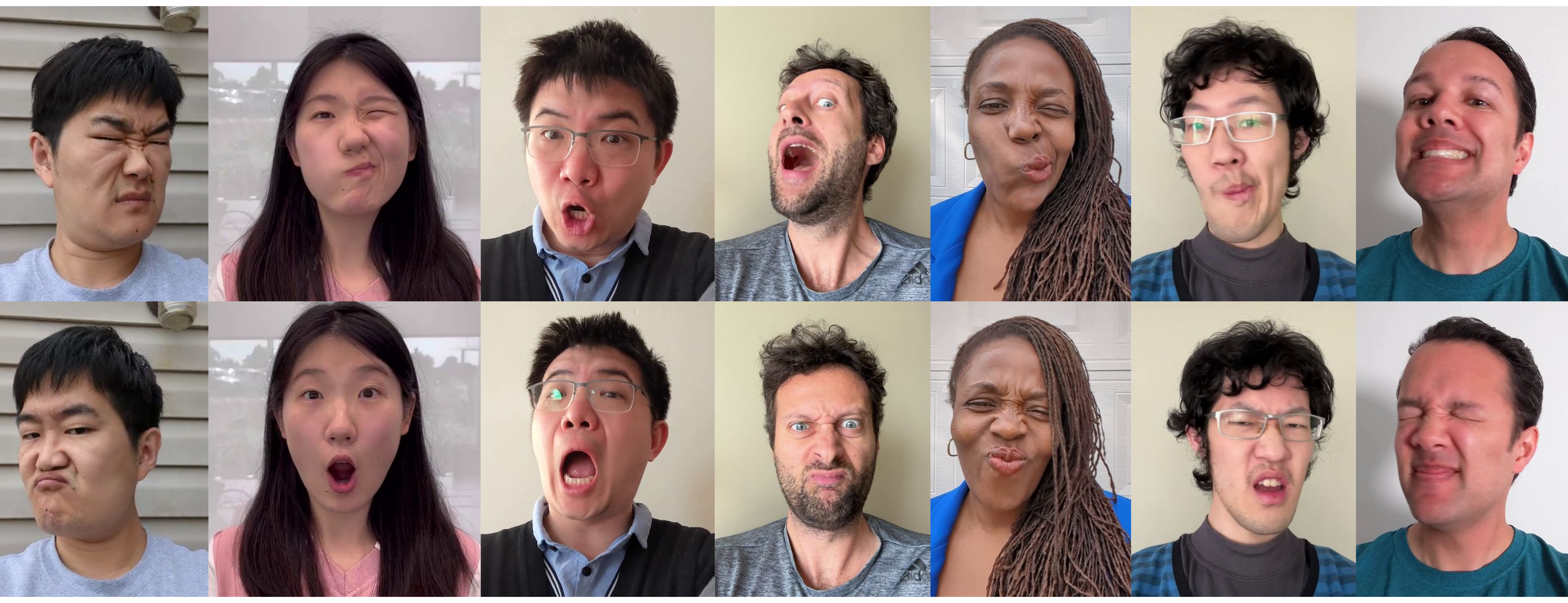}
    \caption{Examples of our captured training data, which includes various large expressions.
    }
    
    \label{fig:data_example}
\end{figure*}

\begin{table}[!htb]
    \setlength\tabcolsep{2.5pt}
    \small
    \centering
      \begin{tabular}{c|c|c|c|c}
        \multicolumn{4}{c|}{Our Data} & NerFACE Data \\ \hline
        \textit{Subject0} & \textit{Subject1} & \textit{Subject2} & \textit{Subject3} & \textit{Subject4} \\
        \hline
         0.657 & 0.610 & 0.589 & 0.796 & 0.426
    \end{tabular}
    \caption{The standard deviations of fitted 3DMM expression codes, averaged across code dimensions, on different subjects.}
    \label{tab:exp_std}
\end{table}

We include data examples (\cref{fig:data_example}) to show that our data has a large expression coverage, thus is more challenging than talking head style data used by prior works \cite{gafni2021dynamic}. We also compare the standard deviations of fitted 3DMM expression codes (averaged across code dimensions) on our data and NerFACE data. As shown in \cref{tab:exp_std}, our data has significantly larger standard deviations, which indicates more diverse expression coverage in our data.


\section{More Results}
\label{sec:more_results}


In this section, we provide more qualitative comparisons of our method with state-of-the-art techniques and ablations against our design choices. We also demonstrate the robustness of our method in challenging cases where the generated avatar is driven under significantly different conditions than the original training sequence.

\subsection{Multi-subject Comparison with SOTA}

\begin{figure*}[t]
    \centering
    \includegraphics[width=\linewidth]{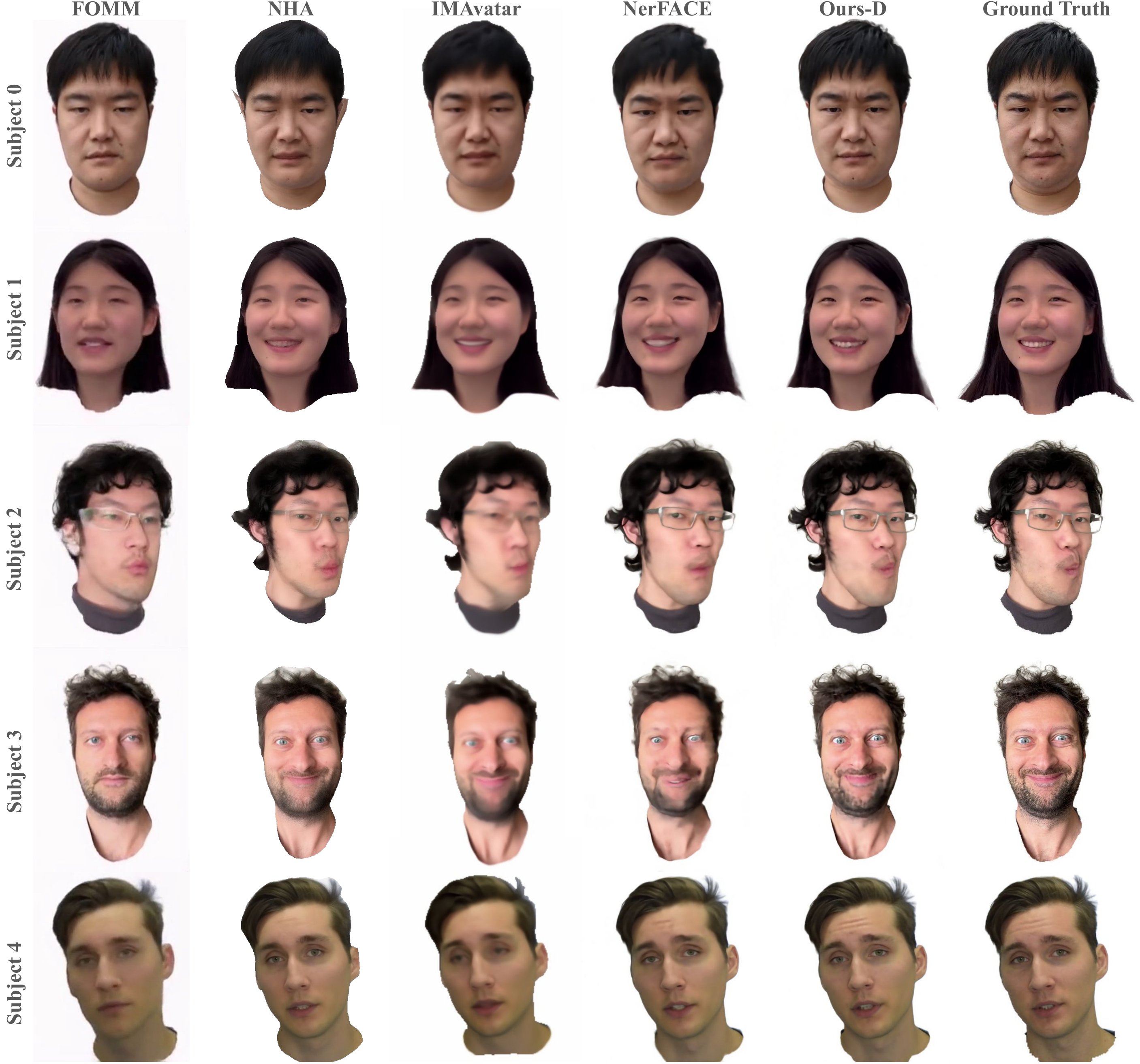}
    \caption{Qualitative Comparison to prior state-of-the-art monocular head avatars. Note how our approach more faithfully reconstructs the ground truth expressions while preserving most of the high frequency details. Please refer to Sec. 4.2 in main paper for more discussions.
    }
    
    \label{fig:vs_sota_1}
\end{figure*}

\cref{fig:vs_sota_1} shows a comparison of our method against state-of-the-art techniques across several subjects for non-neutral expressions. Note that our technique is able to faithfully model and synthesize these challenging expressions across the range of subjects, while preserving fine scale details such as wrinkles and hair, mouth and lip motion, and eye gaze direction, without introducing any significant artifacts. While NerFACE~\cite{gafni2021dynamic} is able to capture the general expression and gaze, it introduces artifacts for example in Subject 3 and produces blurry details on skin and hair due to the limitation of using a single global MLP to model the full appearance. IMAvatar~\cite{zheng2022avatar} and NHA~\cite{grassal2022neural} struggle with capturing volumetric effects in the hair and out-of-model objects such as glasses due to the underlying surface based geometry representation and result in artifacts along the boundaries. FOMM~\cite{siarohin2019first} fails to produce these challenging expressions due to it's inherent 2D representation.

\subsection{Design Ablation Analysis}

\begin{figure*}[t]
    \centering
    \includegraphics[width=\linewidth]{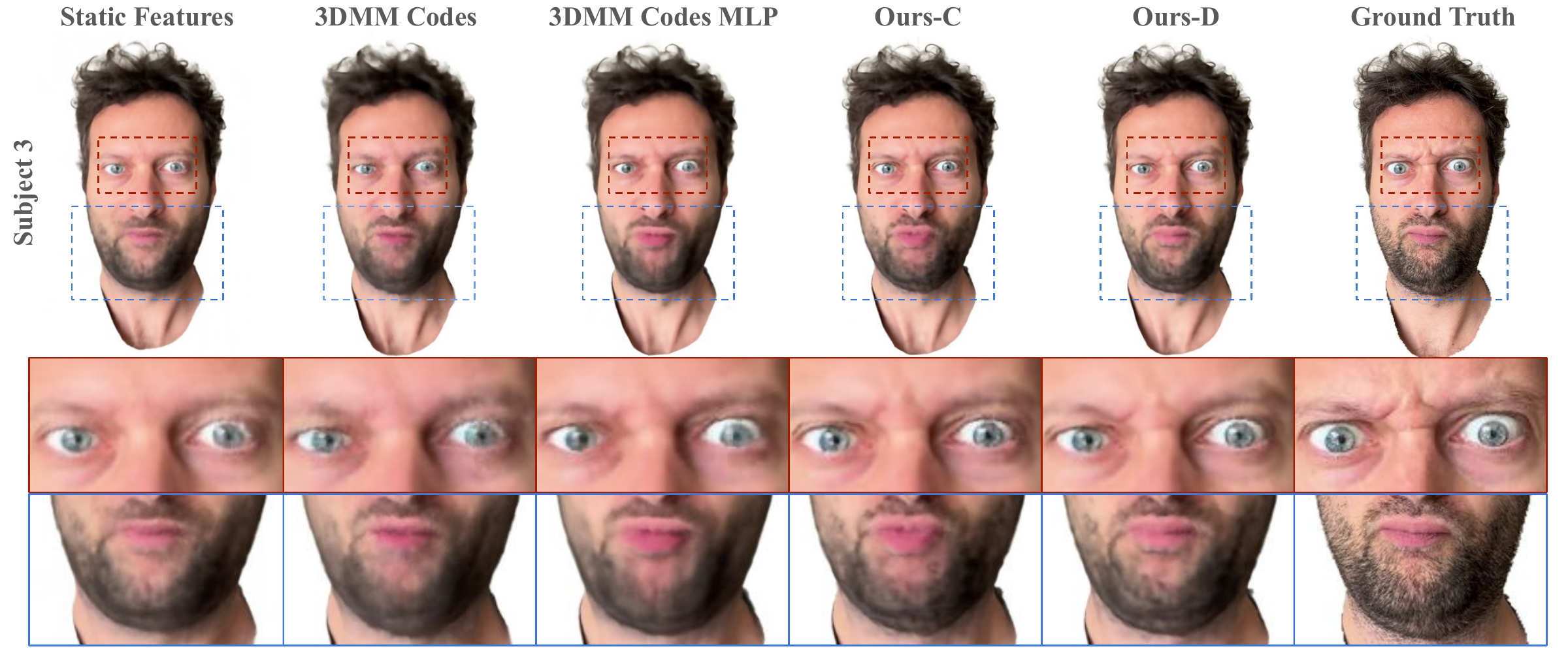}
    \caption{Comparison between different designs for local vertex feature learning. See Sec. 4.4 in main paper for more details. ``Static feature'' struggles to capture personalized expressions. ``3DMM Codes'' improves the personalization but suffers from overall blurriness. ``3DMM Codes MLP'' further improves the sharpness, but still cannot present the details. Overall, our convolution-based methods lead to superior renderings on areas such as eyes, facial hairs, and frown wrinkles.
    }
    \label{fig:ablation_supp}
\end{figure*}

In \cref{fig:ablation_supp} we visualize the close-up result produced by various design choice ablations of our method as detailed in Sec 4.4 of the main paper. These ablations show different ways of predicting per-vertex features on the 3DMM mesh which are spatially interpolated to obtain the volumetric radiance field of the avatar. ``Static Features" learns fixed per-vertex features on the 3DMM mesh over the course of training. Since the features are not conditioned on the expression parameters, it struggles to properly model non-neutral  expressions. ``3DMM codes" concatenates the expression and pose codes to the static features. This results in reproducing the expression better but still results in local artifacts. ``3DMM codes MLP" improves the model capacity by conditioning an MLP based VAE on the 3DMM codes that decodes to vertex features. While this improves the local artifacts, it still produces blurry result due to the global representation. ``Ours-C" uses a convolutional decoder to produce UV space features from 3DMM codes. This significantly improves the level of high-frequency spatial details in the synthesized image. Finally, ``Ours-D" poses the problem as an image-translation task in the UV space by using a convolutional encoder-decoder architecture to directly translate the geometry deformations of the 3DMM to UV space features. This generates local features that achieve the most faithful reconstruction of the expression along with better preserved spatial details.

\subsection{Demonstrating Robustness and Applicability}

\begin{figure}[t]
    \centering
    \includegraphics[width=\linewidth]{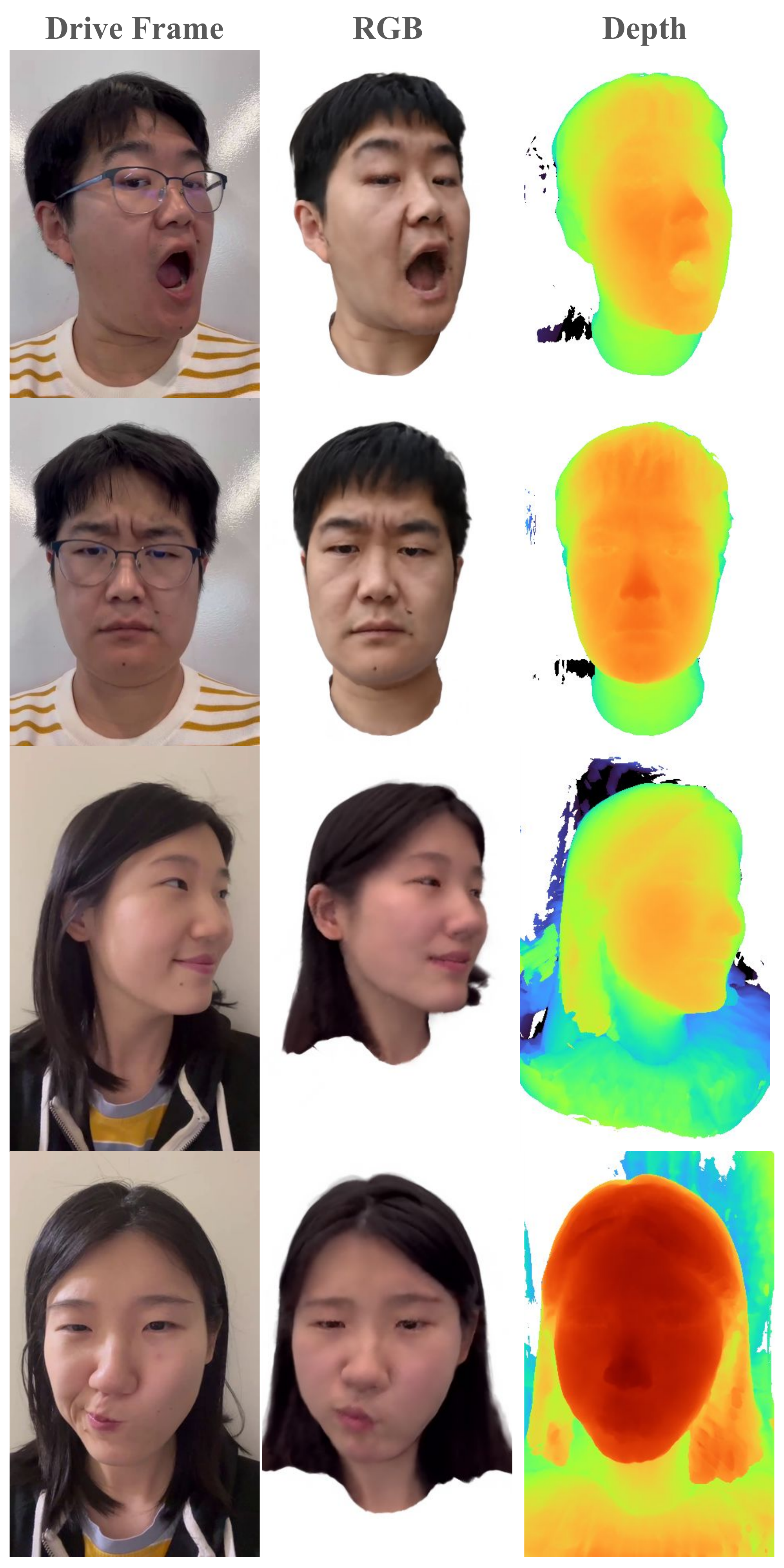}
    \caption{Results on driving the learned avatar by the same subject under different capturing conditions. Our method produces faithful expressions and good geometry.
    }
    
    \label{fig:drive_supp}
\end{figure}

In \cref{fig:drive_supp}, we demonstrate the avatars being driven by the same subject at a different time and place than the original training sequence. Note that the subject's hair style, scene lighting, and accessories such as glasses are different. Our technique is able to faithfully reproduce the pose and expressions even under the novel conditions, demonstrating robustness and practical applicability. Please see the full sequence of this challenging avatar driving in novel conditions in the accompanying supplementary webpage.

\subsection{Video Results}

In the accompanying supplementary webpage, we demonstrate full-sequence results for following cases: 
\begin{itemize}
    \item Driving the avatar using a test clip that is captured in the same conditions as the training data (\ie, same subject, same capturing condition).
    \item Driving the avatar by the same subject under novel conditions of lighting, appearance, and accessories  (\ie, the same subject under different capturing conditions).
\end{itemize}

To drive our avatar, we first obtain camera and 3DMM parameters from the driving video via per-frame 3DMM fitting, then apply these 3DMM parameters to our avatars and render from frontal or novel camera views. Note in the videos that our method produces high-quality controllable avatars that capture identity, pose, and expression specific idiosyncrasies. The avatar can be rendered in 3D from any desired viewpoint. Since the training data is captured only from frontal views, more extreme side views sometimes result in artifacts at the back of the head, which is an expected limitation of our method. Other common challenging cases are people with long hair and wearable. Our method is still able to generate plausible results though indeed shows relatively more artifacts.

Though small temporal jitters are also shown the videos, we observed that the jitters are significantly mitigated when the avatar is driven using synthetically smoothed 3DMM motions. This suggests that the jitters are mainly due to errors in 3DMM fitting. Improved 3DMMs and fitting algorithms in the future would resolve this issue. Future research could also explore the mitigation of temporal jitters from a neural rendering perspective.

\subsection{Visualize 3DMM and Final Geometry}

\begin{figure}[t]
    \centering
    \includegraphics[width=\linewidth]{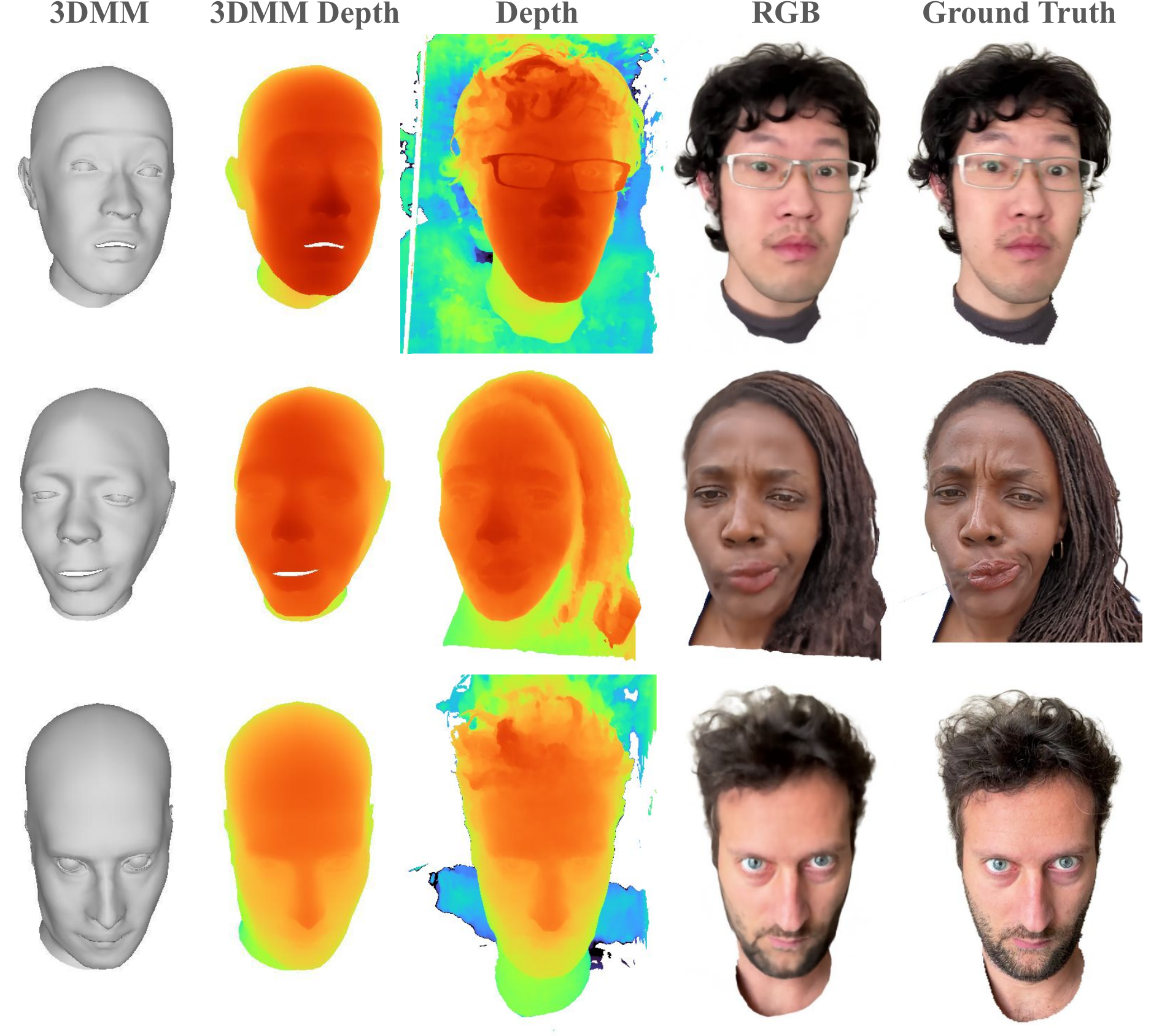}
    \caption{Visualization of 3DMM and final geometry. Our method can reasonably capture out-of-3DMM geometry.}

    \label{fig:3dmm_depth}
\end{figure}

We provide the visualizations and comparisons of the 3DMM mesh and the learned final geometry in \cref{fig:3dmm_depth}. Our method is able to reasonably capture the out-of-3DMM geometry such as glasses and hairs.

\end{document}